\documentclass[11pt]{article} 
\usepackage{ebgaramond}
\usepackage[preprint]{tmlr}
\usepackage{microtype}

\usepackage{amsmath,amsfonts,bm}

\def\eqref#1{equation~\ref{#1}}

\def\1{\bm{1}}

\DeclareMathAlphabet{\mathsfit}{\encodingdefault}{\sfdefault}{m}{sl}
\SetMathAlphabet{\mathsfit}{bold}{\encodingdefault}{\sfdefault}{bx}{n}

\newcommand{\E}{\mathbb{E}}

\usepackage{amsmath}
\usepackage{amsfonts} 
\usepackage{mathtools}
\usepackage{amsthm} 

\usepackage{hyperref}
\usepackage{url}

\usepackage{xcolor}
\usepackage{mdframed}

\definecolor{mygray}{gray}{0.92}
\definecolor{mycolor2}{RGB}{255, 250, 176}

\newmdenv[backgroundcolor=mygray, linewidth=0pt]{sectionpurpose}

\usepackage{xspace}
\newcommand*{\eg}{e.g.\@\xspace}
\newcommand*{\ie}{i.e.\@\xspace}

\newcommand*{\cf}{cf.\@\xspace}
\newcommand*{\versus}{vs.\@\xspace}

\newcommand{\indep}{\perp \!\!\! \perp}

\newcommand{\actfairpred}{\hat{Y}_{\mathrm{af}}}

\newtheorem{theorem}{Theorem}[section]       
\newtheorem{proposition}[theorem]{Proposition}
\newtheorem{definition}[theorem]{Definition}

\title{Insights From Insurance for Fair Machine Learning}

\author{\name Christian Fröhlich \email christian.froehlich@uni-tuebingen.de \\
      \addr Department of Computer Science\\
      University of Tübingen\\ 
      and Tübingen AI Center
      \AND
      \name Robert C. Williamson \email bob.williamson@uni-tuebingen.de \\
      \addr Department of Computer Science\\
      University of Tübingen\\ 
      and Tübingen AI Center
      }

\def\month{MM}  
\def\year{YYYY} 
\def\openreview{\url{https://openreview.net/forum?id=XXXX}} 

\begin{document}

\maketitle

\begin{abstract}
We argue that insurance can act as an analogon for the social situatedness of machine learning systems, hence allowing machine learning scholars to take insights from the rich and interdisciplinary insurance literature. Tracing the interaction of uncertainty, fairness and responsibility in insurance provides a fresh perspective on fairness in machine learning. We link insurance fairness conceptions to their machine learning relatives, and use this bridge to problematize fairness as calibration. 
In this process, we bring to the forefront two themes that have been largely overlooked in the machine learning literature: 
responsibility and aggregate-individual tensions.

%
%

\end{abstract}

\section{Introduction}
Insurance is ``interestingly uninteresting''.\footnote{\citet{mcfall2020personalisation} call insurance ``interestingly uninteresting'', referring to how insurance is ``hugely underresearched'' given its societal importance, which is typically not recognized \citep{ewald1989versicherungs}.} 
In this work, we argue that in fact insurance is far from uninteresting and indeed a rich source of inspiration and insight to scholarship interested in social issues surrounding machine learning, specifically the field now known as fair machine learning. Our proposal is that insurance can be viewed as an analogon to machine learning with respect to these issues arising from the social situatedness. While machine learning is a relatively recent technology, debates regarding social issues in the context of insurance have been ongoing for a long time. Thus, we argue that taking inspiration from studies of insurance can contribute to a more integrative view of machine learning systems as \textit{socio-}technical systems \citep{sociotechnical}. 

Both machine learning and insurance are firmly based on a statistical, probabilistic mode of reasoning --- an \textit{actuarial} mode. Indeed, insurance can be viewed as the first commercial test of probability theory \citep{gigerenzer1989empire,mcfall2011good}. Insurance, a technology for \textit{doing risk}, transforms uncertainty into calculable risk \citep{lehtonen2014editorial}. The key idea is to share the risk of a loss in a collective, organized through an abstract mutuality; due to the `law' of large numbers, uncertainty thus becomes manageable and the effect of chance can be offset \citep{ewald1991insurance}. In this way, insurance creates a ``community of fate'' in the face of uncertainty \citep{heimer1985reactive}. To enter into this community (the insurance \textit{pool}), the insurer demands a certain fee, called \textit{premium}, from the policyholder.

In insurance, questions of fairness inevitably arise, and have been the subject of much debate. The central point of debate is the tension between risk assessment and distribution \citep{abraham1985efficiency}. In other words, who is to be mutualized in the pool. Some form of segmentation is found in many insurantial arrangements: the pool of policyholders can be stratified by separating high and low risk individuals. But the specific nature that such segmentation takes typically depends not only on risk assessment, but on further considerations such as assignment of responsibility, modulated by social context; in this way, insurance is not a neutral technology \citep{baker2002embracing,glenn2003postmodernism}. 

Our non-comprehensive outline of the history of insurance illustrates how uncertainty, fairness and responsibility interact, and can be entangled and disentangled. From this background, we can extract conceptual insights which also apply to machine learning.
The tension between risk assessment and distribution is mirrored in formal fairness principles: \textit{solidarity}, which can be linked to independence in fair machine learning, contrasts with \textit{actuarial fairness}, linked to calibration. Briefly, actuarial fairness demands that each policyholder should pay only for their own risk, that is, mutualization should occur only between individuals with the same `true' risk. In contrast, solidarity calls for equal contribution to the pool.
On one level of this text, we problematize actuarial fairness (by extension, calibration) as a notion of fairness in the normative sense by taking inspiration from insurance. This perspective is aligned with recent proposals that stress the discrepancy of formal algorithmic fairness and ``substantive'' fairness \citep{green2022escaping}, which some prefer to call \textit{justice} \citep{vredenburghfairness}.
Parallel to this runs a distinct textual level, where we emphasize two intricately interacting themes: \textit{responsibility} and \textit{tensions between aggregate and individual}. Both entail criticism of actuarial fairness, but we suggest that they additionally provide much broader, fruitful lessons for machine learning from insurance.

At the highest level of abstraction, our goal is to establish a general conceptual bridge between insurance and machine learning. Traversing this bridge, machine learning scholars can obtain new perspectives on the social situatedness of a probabilistic, statistical technology --- we attempt to offer a new `cognitive toolkit' for thinking about the social situatedness of machine learning. Our point of view is that fairness cannot be reduced to a formal, mathematical issue, but that it requires taking broader social context into account, reasoning for instance about responsibility. And for this, we suggest, insurance is an insightful analogon.  Therefore, our objective is to furnish the reader with a guide that charts the landscape of insurance with respect to social issues and to establish links to machine learning.

On a formal level, we use the following analogy. In a machine learning task, we are given some features $X$ and associated outcomes $Y$, which we attempt to approximate by predictions $\hat{Y}$. The structural relation to insurance is established by conceiving of $X$ as the features of policyholders (\eg age, gender) with outcomes $Y$ (\eg having an accident or not), and the task is to set a corresponding premium $\hat{Y}$.

\section{A Brief History of Insurance Rationalities}
Insurance is not a monolithic technology, but rather a general principle of risk management, which is instantiated in multiple distinct forms. 
Insurance and the conceptual resources it deploys are not immutable and stable over time, as exemplified by \citeauthor{baker1996genealogy}'s \citeyearpar{baker1996genealogy} study on \textit{moral hazard}, attesting to their evolving nature. 
In this section, we provide a succinct, necessarily non-comprehensive description of three historical modes of insurantial operation: the welfare state, neoclassical economics and personalized insurance. 
Throughout, we focus on the role that uncertainty, fairness and responsibility play in each of the three modes.
To each mode we ascribe a set of attitudes towards these three aspects and in what manner they are entangled or disentangled.
Fairness conceptions in insurance are contingent upon prevailing societal norms, particularly regarding responsibility, but concurrently insurance shapes the moral fabric of the society in which it is embedded \citep{glenn2003risk,van2006making,lehtonen2015producing}. Furthermore, fairness conceptions in insurance are historically intertwined with their accompanying (statistical) epistemologies, ways of `knowing' the risk in the face of uncertainty. A common thread is also that distinct forms of insurance correspond to distinct ways of governing society. 
Importantly, we do not want to suggest a linear historic progression here\footnote{See Baker \citeyearpar[p.\@\xspace 571]{baker2000insuring} on Ewald's work for this point.} 
--- different forms of insurance co-exist at any given time. For instance, contemporary health care systems tend to operate with the logic of the welfare state, while actuarial fairness undergirds the private insurance sector. What follows is our synthesis of the literature, particularly drawing on the works of \citet{ewald_letat_1986} and \citet{frezal2020fairness}, with a discerning focus on elucidating the intricate interplay between uncertainty, fairness, and responsibility. With this background, we are then able to extract conceptual lessons that apply also to machine learning.

\subsection{Broad Solidarity in the Welfare State}
In his seminal work \textit{L'Ètat providence}, \citet{ewald_letat_1986} gave an influential account of the rise of the welfare state and explicates how insurance became its prime way of government. The point of departure is the predominance of liberal reasoning, which operates with the categories of \textit{fault} and \textit{foresight} in risk management. Here, it is presumed that the occurrence of an accident (a damage, loss or injury) must be due to fault, a lack of foresight. Liberal thought held individuals responsible for their own fate and inequalities were naturalized as just consequences of individual responsibility \citep{landes2013normative}. An accident implies a trial under the “regime of juridical responsibility” \citep{ewald_letat_1986}, where the goal is trying to establish the fault of one party.
Responsibility is then borne by the person who is assigned fault, who is seen as having caused the accident. In turn, voluntary charity is the preferred means of supporting the poor and unlucky. 
With the rise of industrialization in different countries, an ``accident crisis'' unfolded roughly between 1870 and 1910 \citep{krippner2023unmasked}: the number of workplace accidents dramatically increased and there appeared a new regularity at the aggregate level, suggesting a kind of determinism in the phenomenon. This led to the \textit{objectification} of the accident and its management became a question of the collective, the \textit{social}. By objectification, we understand an act of aggregation combined with the law of large numbers. As a consequence, we find a conception of insurance as broad solidarity and the rise of the welfare states. While at the individual level it was impossible to predict \textit{who} will suffer an accident, the new regularity observed at the aggregate level could provide an effective pattern of risk management. Since equal ignorance in the fate of uncertainty was emphasized \citep{ewald_letat_1986}, 
solidarity, implying equal contribution to the pool, was considered fair. Indeed, this conception of insurance was so firmly based on the aggregate that it led \citet{ewald1991insurance} to assert that 
\begin{quote}
Strictly speaking there is no such thing as an individual risk; otherwise insurance would be no more than a wager. Risk only becomes something calculable when it is spread over a population. The work of the insurer is, precisely, to constitute that population by selecting and dividing risks. [..] It makes each person a part of the whole.
\end{quote}
The business of insurance, then, was construed as the constitution of abstract mutualities\footnote{Typically policyholders are unaware with whom they are mutualized, however, so this mutualization does not require a shared sense of groupness \citep{krippner2022person}.} \citep{lehtonen2011forms} and the sharing of responsibility to counteract the effect of fate.

The epistemology of the welfare state is one of the aggregate, the collective, the social. In this respect, it is interesting and instructive how insurance became intertwined with probability and statistics.
Early forms of insurance were more like \textit{gambling}, and insurance was often accused of being immoral and faced prohibitions.\footnote{\citet{cooper2009probability} provide some examples. In 18th century London life insurance could be bought on the life of celebrities, without an insurable interest to the policyholder. In fact, the notion of an insurable interest was put forward by the insurers to counter the allegation of gambling \citep{baker1996genealogy,mcfall2018or}. Particularly interesting is also insurance in Islamic law, which prohibits gambling and contracts based on usury: the morality of insurance is justified then by emphasizing the solidaristic nature of the arrangement \citep{baker2002risk}, in contrast to the view of insurance as a bilateral contract that is more prevalent in Western societies.} Insurance gained more legitimacy when it became based on `objective' probability and statistics \citep[p.\@\xspace 162ff]{daston2023classical}; by the end of the nineteenth century, the morality of insurance was established \citep{baker1996genealogy}.
The crucial conceptual move was the marriage of probability theory and statistics. While early probabilists were more concerned with reasonable subjective judgment, the nineteenth century shows an increasing shift towards ``objective calculation'' \citep{mcfall2011good}. The idea was that in a large insurance pool losses occur randomly, so the law of large numbers applies and the total loss can be predicted at the level of the aggregate. Frequentist (`objective') probability thus combines aggregate regularity with individual irregularity, as explained by \citet[p.\@\xspace 4]{Venn1876}.


A key player in this development was Adolphe Quetelet, a Belgian astronomer, who initiated the study of ``social physics'' by attempting to discover natural laws about human behaviour \citep{mcfall2011good}.
Quetelet's innovation was the transposition of one sense of the concept \textit{average} to a different one. Consider first the familiar aggregating sense of average, where a set of commensurate objects is summarized in a single number. A prima facie different sense of average is as the single true value of some measurement problem, from which one can obtain a set of noisy measurements. The radical conceptual move was then to transpose the second to the first sense, thereby viewing the individuals of a population as many realizations of some abstract \textit{average human}\footnote{
The original term was ``average man''.
} \citep{ewald1990norms,mcfall2011good}. In this way, the rates of birth, death and other social phenomena could be attributed to this fictional average human.  In the context of the workplace accident and insurance, in line with the imaginary of the average human, the focus shifted from the unique, individual experience (the object of a juridical trial) to an objectification based on the average occurrence \citep{krippner2023unmasked}. With regard to how this influences the notion of personhood, \citet{dean1998risk} writes
\begin{quote}
    Insurance practices displace the abstract, invariant norm of a responsible juridical subject with an individuality relative to other members of an insured population, an `average sociological individuality'.
\end{quote}
Quetelet's fiction of the average human has made an impact on the insurance sector \citep{mcfall2011good}: identifying the individual with an average is at the core of actuarial practice.
Indeed, the question of how individuals relate to the aggregates they make up is, as we suggest in line with \citet{krippner2023unmasked}, runs through the history of insurance. Moreover, we argue in Section~\ref{sec:aggvsind} that it is a major concern for fair machine learning, too.

Although insurance increasingly relied on probability and statistics, the available quantification methods at the level of the collective severely limited the possibility of actually `knowing' the risk of an individual \citep{barry2019rationality}. With improved actuarial methods, new possibilities for segmentation of the pool have opened up, which have led also to the rise of a new fairness notion that is successively contributing to the erosion of solidarity.

\subsection{Neoclassical Economics and Actuarial Fairness}
\label{sec:neoclassical}
A distinct mode of insurance, which undergirds contemporary (private) insurance, is based on neo-classical economics, which construes individuals as rational expected utility maximizers --- the human is viewed as a \textit{homo oeconomicus}. The assumption in this paradigm is that insurance is purchased due to risk aversity from the perspective of the policyholder, while the insurer is risk neutral. 
This configuration of the individual is tied to a different notion of fairness which contrasts starkly with the solidarity of the welfare state: \textit{actuarial fairness}. The idea is that the pure premium (\ie what the policyholder pays before adding additional expenses such as for administration) should equal the expected risk for each policyholder. While the idea of ``equality in risk'' was around for a long time \citep{martinez2016fair}, its modern formulation is due to \citet{arrow1963uncertainty}.
On the one hand, actuarial fairness can be understood as a purely descriptive, technical notion; but it is also advanced in the literature and by insurers as a notion of \textit{fairness} in a normative sense, as legitimate practice, see for instance
\citep{walters1981risk,clifford1987aids,daniels1990insurability,stone2001admissions,thiery2006fairness}; 
Interestingly, actuarial fairness can be traced back to the Aristotelian consistency principle ``fairness is to treat equal people equally and unequal people unequally'' \citep{landes2015fair}, in a context where `equal' means `equal risk' \citep{martinez2016fair}. 

The definition and especially the Aristotelian motivation betrays that actuarial fairness is in practice always a \textit{group-based} notion \citep{miller2009disparate}, since insurers needed (traditionally, at least) to make use of large segments for calculating expected losses. For this, actuaries choose a set of relevant variables, while ignoring others. This seems fundamentally in conflict with the idea of adjustment to the `individual risk' of the policyholder --- and indeed this was not how actuarial fairness was construed until roughly the 1970s, it remained firmly group-based \citep{barry2020insurance}.
In line with Quetelet, \citet{thiery2006fairness} describe the logic succinctly as follows:
\begin{quote}
    [I]nsurance classification schemes rely on the assumption that individuals answer to the average
(stereotypical) characteristics of a group to which they belong. 
\end{quote}

Hence the justification for group-based actuarial fairness relies on assuming an ``average sociological individuality'' \citep{dean1998risk} --- each individual is assigned to a segment and is then identified with the corresponding average \citep{krippner2022person}. In this sense, we find again the logic of the welfare state but now only \textit{segmentwise}, with the aspiration to reduce solidarity between groups as much as possible, given practical constraints.
In the terminology of \citet{lehtonen2011forms}, this means that ideally only \textit{chance solidarity} is left, which compensates for the effects of aleatoric uncertainty;\footnote{We put aside for now the question of whether the conceptual distinction between aleatoric and epistemic uncertainty is well-defined.} in contrast, \textit{risk subsidizing solidarity} refers to a solidarity between individuals of different expected loss.\footnote{\citet{lehtonen2011forms} further mention \textit{subsidizing income solidarity}, occuring when premia are adjusted based on income; this is more like a tax than `genuine' insurance solidarity.}

The role of responsibility in actuarial fairness is subtle. Actuarial fairness, when understood as a normative principle, rests on the assumption that people can be held responsible for their individual risk to some extent \citep{lehtonen2015producing}. However, we should distinguish conceptually between \textit{responsibility} and \textit{responsibilization}, that is, \textit{holding} someone responsible for something.\footnote{We use the term `responsibilization' in line with \citet{andersen2015luck}. The term originally appeared in the governmentality literature and refers to a neoliberal mode of governing that frames individuals as autonomous and responsible, see for example \citep{shamir2008age,pyysiainen2017neoliberal}.} While normative philosophical literature may be careful about this distinction, in practice it can appear blurry. In fact, there are two principles that provide \textit{non-responsibility based reasons for responsibilization}  in insurance \citep{andersen2015luck}, therefore in favor of actuarial fairness: \textit{moral hazard} and \textit{adverse selection} (see Appendix~\ref{sec:moralhazardadverse}).
Thus, responsibility is, in the neoclassical framework, not central to actuarial fairness \citep{landes2015fair}.
However, the role of responsibility (more precisely, responsibility-based reasons for responsibilization) in insurance is currently being emphasized more and more. To this we turn now.

\subsection{The Climax: Personalized Insurance}
\label{sec:personalized}
From the 1980s on, the insurance industry was increasingly challenged by anti-discrimination legislation. Social movements attacked the average human (woman) logic that insurance based its actuarially `fair' premia on. A prominent example is the campaign initiated by the National Organization for Women (NOW) \citep{krippner2022person,krippner2023unmasked}, aimed at ending gender-based risk segmentation. In line with the civil rights movement, feminists considered such underwriting (that is, risk classification) practices to be unfairly discriminatory, as they rely on group-based generalizations. Instead, they asked for a finer adjustment to \textit{individual risk}. What was under attack here is the fundamental group-based logic of actuarial fairness. The US supreme court asserted in the context of insurance that ``[e]ven a true generalization about [a] class cannot justify
class-based treatment'' (Norris, 1983, as cited in \citet{avraham2017discrimination}). A more recent case is the \textit{ECJ Test-Achats} ruling in the EU, which highly restricts gender-based underwriting \citep{rebert2015right,cevolini2022actuarial}. 
In response to such anti-discrimination legislation and in anticipation of a continuation of this trend, the practical meaning of actuarial fairness gradually began to shift.
Consequently we find two highly entangled trends in the contemporary insurance industry: the \textit{individualization of risk} and \textit{behaviour-based personalization} \citep{cevolini2020pool,mcfall2020personalisation}.

The individualization of risk can be understood as taking group-based actuarial fairness to the limit and (hypothetically) forming `groups of one'; instead of spreading risk over a pool, each policyholder would pay exactly for her own \textit{individual risk} \citep{cevolini2020pool}. In terms of solidarity, this implies a complete erosion of subsidizing risk solidarity, so that ideally only chance solidarity remains. Rephrasing this, what is now being emphasized is the non-homogeneity within previous groups of policyholders \citep{barry2020insurance}.
To this end, insurers begin to shift the focus from attributes which are considered uncontrollable (\eg gender, race)\footnote{While the case of gender demonstrates that what is considered controllable can change, for insurance purposes, gender is arguably still uncontrollable.} to controllable, dynamic data about the individual, and adjust premia accordingly, yielding behaviour-based personalization. Of course, the hope is that behaviour is closely linked to the individual risk, otherwise personalization would hardly be reasonable; in this way individualization and personalization are correlated.\footnote{One would (in most contexts) not try to personalize premia based on the binary feature `having attached earlobes'.} 

Personalization is linked to \textit{InsurTech}, that is, technology-driven innovation in insurance \citep{mcfall2020personalisation}. A prominent example is the use of wearable devices such as fitness trackers in health insurance \citep{lupton2016diverse,mcfall2019personalizing}, where discounts and rewards are supposed to incentivize ``healthy behaviour''. In car insurance, the use of telematics is gaining popularity \citep{verbelen2018unravelling,meyers2018enacting,cevolini2022actuarial}, where a small device installed in the car dynamically provides information about driving behaviour from proxy variables such as speed to the insurer, as well as feedback to the policyholder. Here, the premium is continuously adjusted to behaviour, which contrasts with previous static underwriting. In both examples, the premium is supposed to act \textit{on} the behaviour with the aim of loss prevention \citep{mcfall2018or}; this opens up the possibility for feedback loops. In other words, the premium is \textit{performative} --- see Appendix~\ref{sec:performativity}.

The accompanying epistemology associated with the shift towards individualization and personalization is the aim to tailor the premium to the `individual risk' 
and it is assumed that a combination of big data (often behavioural, with a fine temporal resolution) and machine learning enables `knowing' this risk \citep{cevolini2020pool}. Hence, in this currently unfolding chapter, machine learning enters into a dynamic interaction with insurance; we expect conceptual lessons to flow in both directions in the future (compare also \citep{williamson2004dynamic}); in this paper, however, we specifically focus on lessons from insurance for machine learning. \citet{barry2019rationality} describes the new epistemology as follows:
\begin{quote}
    Hence what was once considered as `noise,' the
individual specificities that had to be averaged out by statistics, is now the core of the
analysis and the focus of the new knowledge.
\end{quote}
The upshot, according to \citet{barry2019rationality}, is the ``deconstruction of the aggregate viewpoint that produced collectives''. Commentators speak of emerging ``segments of one'' \citep{prainsack2020shifting}. 

The individualization and personalization of risk is associated with a shift in fairness: actuarial fairness is taken to the limit and now clearly carries a normative flavour based on a linkage to  \textit{responsibility}. We call this utopia of individual risk adjustment \textit{perfect actuarial fairness}, to demarcate it from practical, group-based actuarial fairness. Reviving pre-Welfare liberal thought, individual responsibility is stressed \citep{dean1998risk}. For example, \citet{ericson2000moral} document a shift in the concept of accident; the new rhetoric, speaking of a ``crash'' in the case of a car accident, underscores that \textit{someone must be at fault} and thus responsible. Without taking a philosophical stance on actual responsibility, the trend is one of the \textit{responsibilization} of the individual. Even in the context of health insurance, individuals face such responsibilization \citep{van2006making,prainsack2020shifting}. For example, self-tracking favors a view of individuals as ``managers'' of their health \citep{lupton2016diverse,sharon2017self}. Overall, responsibilization is linked to neoliberal modes of government \citep{dean1998risk,ericson2000moral,meyers2018enacting}.

As a consequence, many commentators argue that personalization undermines solidarity \citep{Rosanvallon2000-ROSTNS-2,prainsack2020shifting,barry2020insurance,cevolini2020pool}; For instance, \citet{swedloff2014risk} claims that big data is in contradiction to the risk-spreading mechanism of insurance and \citet{martinez2016fair} observe that, when taken to the extreme, actuarial fairness contradicts the very logic of insurance.\footnote{Some argue that there is no place for risk subsidizing solidarity in insurance, that insurance is concerned only with chance solidarity. However, the conceptual distinction between these forms of solidarity is unclear and rather heuristic \citep{frezal2020fairness}; \cf also the discussion in Section~\ref{sec:aggvsind} on individual risk.} When risk spreading disappears, insurance becomes more like personal saving. With respect to distributive consequences, the individualization of risk has been ``profoundly inegalitarian'' \citep{armstrong2005equality}. The highest-at-risk individuals can even face exclusion from the pool \citep{lehtonen2015producing,cevolini2020pool}.

In summary, we have described three broad modes of insurance and their associated attitudes towards uncertainty, fairness and responsibility. We now investigate multiple dimensions of responsibility and then establish a link to fair machine learning, where we argue that reflections on responsibility should be foregrounded.

\section{Responsibility}
Insurance actively constructs and distributes responsibility \citep{baker2002risk}; the boundaries of individual responsibility are drawn by accounting for some factors but not for others in the premium. 
Adding to the terminology of \citet{landes2015insurance}, we distinguish four dimensions of responsibility: causal (\textit{who is causally responsible for the accident?}), control-based (\textit{who could control the happening?}), moral (\textit{who is normatively responsible?}) and material (\textit{who bears the consequences?}). By responsibilization we understand holding individuals responsible, with an emphasis on the material dimension, based on narratives about causal, control-based and moral dimensions of some phenomenon. What is distinctive about insurance as a technology of risk management is the tendency to separate these dimensions \citep{landes2015insurance}. However, we have observed a recent trend towards a renewed entanglement when compared to the mode of the welfare state. In particular, the notion of actuarial fairness has been increasingly linked to responsibility in contrast to mere non-responsibility based responsibilization.

What is intriguing is also how the entanglement of uncertainty and responsibility has changed historically. A thick veil of ignorance, when the individual level remains out of reach, appears to favor a collective responsibility for risk management; when this veil is gradually lifted, it seems easier to assign responsibility to individuals \citep{frezal2020fairness,barry2020insurance}. However, this is not a necessity: `knowing' individual risk (to some extent) does not necessarily imply that individuals are morally responsible or that we should hold them materially responsible (see Section~\ref{sec:contingencies}).


\subsection{Causality and Control}
\label{sec:causalitycontrol}
Causality has received major attention in recent machine learning research and is also widely discussed in the literature on insurance. A highly related, but subtly distinct notion is that of control --- indeed, the recent trend of responsibilization can to some extent be explained as a response to legislation that prohibits differentiating premia based on variables beyond individual control \citep{meyers2018enacting}.
While actuarial practice is correlation-based, causality and control are emphasized by legal commentators on insurance discrimination (see \eg \citep{abraham1985efficiency,gaulding1994race,avraham2013understanding,avraham2017discrimination}). For example, \citet{avraham2017discrimination} demands that a variable used for calculating premia must be both causally linked to the outcome of interest \textit{and} within individual control. Why demand a causal relationship? If a variable is merely non-causally correlated with the outcome, then it is a proxy for another variable, which should be used in its place in order to avoid differential inaccuracy \citep{avraham2013understanding}.

Another argument is that the importance of causality is derived from control, and that it is control which is at the heart of many controversies. To fix a rough notion of causality, we understand ``$X$ causes $Y$'' as ``intervening on $X$ changes the probability for $Y$'', where an intervention means changing the value. In this way, we can view causality as hypothetical control. Further, if an individual can \textit{actually} intervene on $X$, then we may say simply that $X$ is under control of the individual.\footnote{As an example, genes might be causally related to some outcome $Y$ of interest (hypothetically controlling genes would change the probability for $Y$), but are not actually under control of the individual.} Observe that ``$X$ causes $Y$'' is thus a necessary but not sufficient condition for an individual's ability to control $Y$ by controlling $X$.
The importance of control in turn derives from responsibility: how could an individual be responsible (in the moral and perhaps in the material sense) for a variable which is beyond control \citep{abraham1985efficiency,avraham2017discrimination}? Mere causality seems insufficient for this. Hence, we view causality as the conceptual entry point to get at control. Moreover, control is key for downstream effects, as we discuss in Appendix~\ref{sec:performativity}. From a normative perspective on responsibility, the importance of control (or more precisely, choice) is emphasized in theories of \textit{luck egalitarianism}, often discussed as a justification of risk classification \citep{knight2013luck,lippert2015genetic,huseby2016can,bjork2020better}.

Conversely, control is arguably not \textit{sufficient} for responsibility. 
For instance, control without causality can yield `discrimination by proxy': in \citeauthor{swedloff2014risk}'s \citeyearpar{swedloff2014risk} hypothetical example, liking Vampire novels (arguably within control) is correlated with risky behaviour, but in a non-causal way. This suggests that here a controllable variable works as a proxy for a potentially non-controllable one such as gender. To further complicate matters, the argument by \citet{hu2020s} demonstrates that a conflict may arise when a seemingly controllable stands to a non-controllable variable such as gender in a \textit{constitutive} relation;\footnote{For constitutive relations, see also the discussion on performativity in Appendix~\ref{sec:performativity}.} responsibilizing for the controllable variable then effectively leads to responsibilization for the non-controllable one, too.

In the context of (fair) machine learning (see Section~\ref{sec:fairmllink}), causality has received much attention, but due to the previous considerations we suggest putting reflections on control-based responsibility at the center. This also implies shifting the focus to downstream (`performative') effects of deploying a machine learning system, since consequences of responsibilization are linked to control (see Appendix~\ref{sec:performativity}). Problematically, however, we must answer the question of what is under control. We now argue, in line with other social studies of insurance scholars \citep{abraham1985efficiency,gaulding1994race}, that the variant of this question which is relevant for insurance and machine learning purposes is in fact fundamentally \textit{normative} in character.

\subsection{Social Contingencies in Responsibilization: What Is Under Control?}
\label{sec:contingencies}
The distinction of \textit{control vs. no-control} plays a major role in justifying the responsibilization of individuals, and thus actuarial fairness in the mode of personalization.\footnote{Recall that in the neoclassical mode, the question of control is disregarded, but it is emphasized in the mode of personalization.} Here we illustrate with examples from insurance that this dichotomy is a slippery one, however, and thus provides only a shaky basis for actuarial fairness. The question of whether a variable is within individual control is often unclear and in fact insurance scholars have argued that it is a normative question \citep{abraham1985efficiency,gaulding1994race}, not a descriptive one. Our sketch of an argument is as follows.

An instructive example is risk classification based on smoking in life insurance. It was only in the 1980s that life insurance companies widely started to differentiate premia with respect to smoking, even though it was already known in the 1960s that the associated difference in lifespan is substantial --- the decision \textit{not} to responsibilize depended on the social acceptability of smoking \citep{wilkie1997mutuality,glenn2003risk}. Today, `lifestyle'-based responsibilization increasingly plays a role in insurance and smoking is considered a prime example for a variable within individual control \citep{van2006making,van2007genetic}. We should thus pay close attention to shifting societal narratives about control and responsibility.
A contrasting example is the case of HIV in underwriting practices, which \citet{daniels1990insurability} explicitly contrasts with the (previous) neglect of smoking as a rating variable. \citet{daniels1990insurability} discusses a widespread denial of health insurance coverage for individuals with HIV with a justification based on actuarial fairness. However, \citet{daniels1990insurability} suspects that homophobia and social antipathy to drug users may play a role in this, favoring responsibilization for risky `lifestyle', conceived as controllable.

Underwriting based on `lifestyle' risks can be most starkly contrasted with the use of genetic information in health and life insurance, which is now tightly regulated in some countries \citep{van2018genomics} albeit welcomed by the industry \citep{rechfeld2016personalised}. The notion that ``we are all carriers of genes'' has successfully invoked a solidaristic imaginary in this context \citep{van2007genetic}: many argue that nobody should be penalized for their genes, since they are clearly beyond individual control. Here, a `genetic veil of ignorance' is mobilized in the debate. This has even given rise to the paradigm of \textit{genetic exceptionalism}, holding that genetic information is normatively distinct from other medical information (for a critique see \citep{lippert2015genetic}). In the context of genetic information, solidarity is thus emphasized \citep{liukko2010genetic}, but this has on the other hand contributed to a responsibilization of `lifestyle' risk which continues to justify actuarial fairness \citep{van2018genomics}. 

Do the previous cases really show that the question of control is a normative one? We are not opposed to the idea that there exists a prior, descriptive question of control: considering \textit{all} possible actions\footnote{Even granting that such an expression is sensible.} that an individual can embark on, does one of them intervene on $X$? In this way, it seems reasonable to say that for instance driving behaviour, but not genes, are controllable. This, however, misses the point as it is not the relevant question in the context of responsibility. Control-relevant actions will bring about different consequences for the individual, and what is more, those consequences will differ among individuals. To give up smoking might give more negative utility to an individual with genetic dispositions that favor addictive behaviour. To move to a less earthquake-prone area, in order to lower one's insurance premium, might require investing a large amount of one's resources. Reasoning about the control dimension of responsibility then amounts to setting the boundaries of which actions we may justifiably demand from an individual, and hence becomes normative in character, in effect a matter of distributive justice.
Acknowledging the normative element in questions of control offers a new lens on the \textit{fairness} of actuarial fairness, demonstrating that actuarial calculations are not as `objective and neutral' as they are promoted, a point which we further develop in Section~\ref{sec:performativity}.
While our examples are from insurance, we want to transport conceptual insights to machine learning.

\section{The Link to Fair Machine Learning}
\label{sec:fairmllink}
In recent years, as machine learning is increasingly being deployed in sensitive domains, the field of \textit{fair machine learning} has flourished \citep{barocas-hardt-narayanan,mitchell2021algorithmic,mehrabi2021survey,castelnovo2022clarification}. Different mathematical formalizations of fairness have been proposed in the literature (see \eg \citep{barocas-hardt-narayanan}); we here focus on statistical, group-based definitions. 
As is common in the literature we fix a probability space, for simplicity assumed finite, where we define the following random variables: $X \in \mathcal{X}$ represents the features of the individuals under consideration, $Y$ represents the true outcomes associated with those individuals, $\hat{Y}$ represents the predictions generated by our model, and $S \in \mathcal{S}$ represents a `sensitive feature'\footnote{In the fair machine learning literature, a sensitive feature relates to membership in a socially salient group, for instance based on gender, race or religion.} related to the individuals. We assume that $S$ can be perfectly predicted from $X$, \eg $X=(\tilde{X},S)$ for some $\tilde{X}$.
For simplicity, we assume binary $Y \in \{0,1\}$ and probabilistic scores $\hat{Y} \in [0,1]$. For example, in a credit lending scenario, $X$ contains features such as age, income etc., $S$ could represent ``having migrant background'' in a binary way,
$Y$ indicates whether an individual defaulted or not and $\hat{Y} \in [0,1]$ represents the probabilistic prediction of the model. Group-fairness definitions are often based on binary decisions, but for the analogy with insurance we use probabilistic scores: we intuitively think of $Y$ as representing the true outcome (an accident, damage or loss) and $\hat{Y}$ as representing the premium that the insurer (by analogy, the ML engineer) demands for shouldering the risk of the uncertain outcome $Y$. By $\indep$ we denote statistical independence of random variables. 

\begin{definition}
\label{def:independence}
    A model satisfies \emph{independence} if \thinspace\thinspace $\hat{Y} \indep S $.
\end{definition}
If, for instance, $S$ represents migrant background, independence demands that the distribution of scores is the same for people with and without migrant background. Independence starkly contrasts with calibration.


\begin{definition}
\label{def:calibrationwithingroups}
    A model satisfies \emph{calibration by groups} with respect to $S$ if 
    \begin{displaymath}
        \E[Y \mid S=s, \hat{Y}=\hat{y}] = \hat{y}, \quad \forall s \in \mathcal{S} \thinspace \thinspace \forall \hat{y} \in [0,1].
    \end{displaymath}
\end{definition}

As a fairness criterion, calibration embodies the aim of matching our probabilistic predictions well to the true outcomes. If our model predicts a probability of $p$\% for defaulting, then indeed $p$\% should default if our model is adequate. 
Recently, \citet{holtgen2023richness} have demonstrated that in fact calibration is a richer notion than what is captured by the traditional definition. While they focus on the case of finite data, we present a corresponding theoretical definition. Assume again some set of groups $\mathcal{G} \subseteq 2^{\mathcal{X}}$. Calibration can then be defined based on this choice of groups.
\begin{definition}
\label{def:calibrationtheo}
A model satisfies (theoretical) calibration with respect to a set of groups $\mathcal{G}$ if:
    \begin{displaymath}
\forall G \in \mathcal{G}: \E[(\hat{Y}(X)-Y)|X \in G]=0.
\end{displaymath}
\end{definition}
While this looks deceptively similar to the recently popularized \textit{multi-calibration}, it abstracts away from the prediction-based binning, which is partly due to historical reasons \citep{holtgen2023richness}.
To gain intuition, it is instructive to consider what this means in the case of finite data.
Assume a finite dataset $(X_1,Y_1),..,(X_n,Y_n)$ with associated predictions $\hat{Y}(X_1),..,\hat{Y}(X_n)$. 
A set of groups is then equivalent to choosing a subset of the data, \ie a set $\mathcal{G} \subseteq 2^{\{1,..,n\}}$.
If we use the empirical distribution associated with this dataset in Definition~\ref{def:calibrationtheo}, we obtain the following empirical variant.
\begin{definition}
\label{def:generalcalibration}
    A model satisfies (empirical) calibration with respect to a set of groups $\mathcal{G}$ if:
    \begin{displaymath}
         \forall G \in \mathcal{G}: \frac{1}{|G|} \sum_{i \in G} (\hat{Y}(X_i) - Y_i) = 0.
    \end{displaymath}
\end{definition}
This offers an insurantial interpretation: $\hat{Y}(X_i)$ is the premium we demand for shouldering the risk of the uncertain $Y_i$. Indeed, calibration is formally reminiscent of the subjective betting interpretation for probability theory proposed by \citet{de2017theory}.
There, the expectation $\E[Y]$ is viewed as the fair betting price for a gamble $Y$.\footnote{Accordingly, probability refers to indicator gambles, that is, indicator functions of events.} Calibration is however not tied to a subjective or objective interpretation of probability, but a criterion for model evaluation.

Actuarial fairness and calibration are in close correspondence. For a given set of groups $\mathcal{G} \subset 2^{\mathcal{X}}$, which we assume forms a partition of $\mathcal{X}$, we define the actuarially fair predictor $\actfairpred(x) \coloneqq \E[Y|X \in G_x]$, where $G_x$ is the unique group that contains $x$. The goal of actuarial fairness is to make this partition as fine as possible (\cf Section~\ref{sec:neoclassical},~\ref{sec:personalized}).

  Consider first the extreme choice of a single group as the whole population in Definition~\ref{def:generalcalibration}. 
  Then calibration in the theoretical (respectively, empirical) case demands that  
\begin{displaymath}
\E[\hat{Y}(X) - Y]=0, 
\quad \frac{1}{n} \sum_{i=1}^n (\hat{Y}(X_i) - Y_i)=0,
\end{displaymath}
which is called \textit{global balance} in insurance \citep{denuit2021autocalibration}; intuitively, we need to collect sufficient premia $\hat{Y}(X_i)$ to cover all claims $Y_i$. In this case of a single group, calling $\actfairpred$ `actuarially fair' is some abuse of naming, since in this case it corresponds to full solidarity.
\begin{proposition}
\label{prop:actfair}
    Given a partition $\mathcal{G} \subset 2^{\mathcal{X}}$ of $\mathcal{X}$, the actuarially fair predictor $\actfairpred$ satisfies (theoretical) calibration with respect to $\mathcal{G}$, and is furthermore the coarsest calibrated predictor in the sense that any other predictor which is calibrated with respect to $\mathcal{G}$ either coincides with it or is not group-wise constant.
\end{proposition}
The trivial proof is in Appendix~\ref{app:actfairproof}.
If we had access to the true outcomes $Y$, we could use them for the finest, calibrated predictions. In this way, calibration is still a coarser criterion as it is based on the expectation, the theoretical average, and thus aligned with actuarial fairness; perfect accuracy is in general not demanded.
In the extreme, making the partition finer and finer we reach segments of one. Calibration then demands that $\hat{Y}(x) = \E[Y|X=x]$, which we called perfect actuarial fairness.
In this way, we obtain a `spectrum of calibration', where refining the choice of groups interpolates between two extremes. Group-based actuarial fairness attempts to approximate the extreme of segments of one given practical constraints. Hence, actuarial fairness is well-aligned with fairness-unaware machine learning, where the goal is to approximate the conditional expectation $\E[Y|X]$ as closely as possible. In the limit, the distinction between group-based approaches to fairness and \textit{individual fairness} (in the sense of the machine learning literature \citep{dwork2012fairness}) then becomes blurry: perfect actuarial fairness corresponds to the notion of \textit{individual merit} \citep{joseph2016fairness}, but in line with \citet{binns2020apparent} we argue in Section~\ref{sec:aggvsind} that this does not yield actually `individual' fairness.

Actuarial fairness is closely related to calibration due to Proposition~\ref{prop:actfair}; however finer predictors (\eg the perfect predictor) satisfy calibration, as well. For a precise conceptual correspondence with perfect actuarial fairness, we could consider the following class of fairness measures, inspired and slightly generalized from \citet{raz2021group}.\footnote{\citet{raz2021group} defines a fairness measure as conservative if it is necessarily satisfied by the perfect predictor $\hat{Y}=Y$.}
\begin{definition}
    A fairness measure is probabilistically conservative if it is necessarily satisfied by the perfect actuarially fair predictor $\actfairpred(x)=\E[Y|X=x]$.
\end{definition}
For simplicity and due to the insurantial interpretation, we focus on calibration.

Not only for calibration, but similarly for independence (Definition~\ref{def:independence}) the choice of grouping is crucial. At the one extreme, choosing $S$ to be the indicator of the whole population, independence becomes vacuous as it is trivially satisfied. In contrast, we can add more and more groups (sensitive features) for which we demand independence, so that less and less variations in $\hat{Y}$ are allowed. In the limit, then, we reach \textit{full solidarity} with a constant $\hat{Y}$ (see Appendix~\ref{app:solidarityproof}).
Contrasting independence and calibration, the question is whether we allow the prediction $\hat{Y}$ (the premium) to be sensitive to a certain group or not --- where the within-group variation is however neglected. 

Calibration and independence can be mapped onto \textit{responsibilization vs. non-responsibilization}. The fairness notion that is embodied by calibration is that of accurately reflecting the `true' probabilities, that is, holding individuals responsible for their risk.
In contrast, independence works to decouple predictions from `true' probabilities to some extent and thus can be viewed as non-responsibilizing --- independence is similar, albeit not equivalent, to affirmative action \citep{raz2021group}. As a consequence, we expect different \textit{performative effects} (see Appendix~\ref{sec:performativity}), \ie downstream effects, when applying calibration vs. independence. 

For practice, a simple suggestion is as follows. Since calibration comes practically for free for loss-minimizing predictive models \citep[p.\@\xspace 62f]{barocas-hardt-narayanan}, we may focus on demanding certain independence relationships. Assume that we have designated a subset of features $X_R$ which we aim to responsibilize for, and a set of features $X_{NR}$ which we aim \textit{not} to responsibilize for. Hence $X=(X_R,X_{NR},X_{\text{other}})$. Conditional independence \citep{castelnovo2022clarification} then demands that 
\begin{displaymath}
    \hat{Y} \indep X_{NR} \mid X_R
\end{displaymath}
The features $X_{\text{other}}$, on which we withhold judgement, can then be used by the model in a way restricted by the conditional independence.
In the spirit of Section~\ref{sec:contingencies}, however, the choice of features for (non)responsibilization should be the outcome of a reflexive process of inquiry.

Beyond calibration and independence, other proposals have been put forward in the machine learning literature, which may also be linked to (non)responsibilization; hence the lessons from insurance can be applied, too. For instance, within an equality of opportunity framework, \citet{heidari2019moral} suggest splitting the whole set of features into a set of ``accountability'' features and ``irrelevant'' features. From our perspective, this maps onto responsibilization and non-responsibilization. As another prominent example, in a causal fairness framework, \citet{kilbertus2017avoiding} assume that a set of ``resolving variables'' is given, which are influenced by a sensitive feature in a way that it considered ``non-discriminatory''; but the authors do not provide guidance on how to choose them. 
We believe that the lessons from insurance about causality and control (Section~\ref{sec:causalitycontrol}), and more broadly on responsibility in general, can on the one hand guide the selection of such features. On the other hand, we have seen that causal and control-based dimensions of responsibility are highly sensitive to social context (Section~\ref{sec:contingencies}). This highlights the normative element in making such a distinction (responsibilizing or not), which can be problematic and must be recognized as such. The choice can be side stepped by favoring solidarity over actuarial fairness.



\section{Tensions between Aggregate and Individuals}
\label{sec:aggvsind}

Throughout the history of insurance, and also highly relevant to machine learning, we find tensions between \textit{aggregate and individual}. The mode of the welfare states operates with the imaginary of a collective, in which the individual is mutualized in solidarity. This aggregate viewpoint, where an individual is always identified with the average of some group, finds continuity in group-based actuarial fairness \citep{thiery2006fairness} --- consistent with Quetelet's \textit{average human}. Social movements, however, argued that the group-based actuarially fair price is not fair from the viewpoint of the individiual --- the desire was to ``navigate the social world \textit{unmarked} by the social stereotypes (fashioned by actuarial science as `objective' statistical classifications) [..]'' \citep[emphasis in original]{krippner2023unmasked}. This critique has prompted a shift in the enactment of actuarial fairness \citep{meyers2018enacting}, giving rise to the individualization and personalization of risk and thereby threatening (risk subsidizing) solidarity by dissolving the aggregate. 
At the heart of this aggregate \versus individual tension is the multifaceted concept of responsibility: what links the individual to the aggregate is mutualization based on establishing shared responsibility. \textit{Personhood}, being an individual, is deeply intertwined with assigning responsibility, as insurance scholars have pointed out \citep{mcfall2018or,moor2018price}. It is never the whole of a person that a premium is attached to in insurance, but specific, contextually relevant aspects \citep{mcfall2018or}, thereby transforming a person into an insurance risk \citep{van2014politics,mcfall2018or}.
It would be intriguing to explore how personhood is negotiated within machine learning systems, drawing parallels with similar studies in insurance (\eg \citep{tanninen2020contested}).

The mode of machine learning is a paradoxical one: on the one hand, it fits with the mode of individualization and personalization. The goal is to provide highly tailored predictions for the individual. On the other hand, aggregates are central to the workings of machine learning: they appear in the input data due to categorization processes; second, the fairness of machine learning systems is typically evaluated based on groups (with the exception of \textit{individual fairness}, see below); third, machine learning in general, whether fairness-unaware or not, rests on aggregate criteria such as average training error. We suggest that a large share of the social worries and issues surrounding machine learning can be understood by framing them in the context of the aggregate \versus individual tension.

Group-based actuarial fairness, which relies on historical data aggregated by groups, is prone to reproduce past injustice \citep{daniels1990insurability,lehtonen2015producing}, see also  Appendix~\ref{sec:performativity}. In contrast, the allure of perfect actuarial fairness associated with the personalization of risk, driven by big data and machine learning, is that it is supposedly \textit{individually fair} --- the goal being `segments of one' and setting the premium as $\E[Y|X=x]$. However, we contend that this elusive goal cannot be reached.
The core issue lies with the `hidden collective'. The working of a neural network is similarity-based computation, arguably interpolation \citep{hasson2020direct}. Predictions are invariably grounded in data from individuals \textit{similar to you}, where the similarity is with respect to the opaque nonlinear character of the network. This argument has been made in the context of insurance: `individualized' risk is still relative to the other members of the collective \citep{tanninen2020contested,prainsack2020shifting}. Yet individual justice in an Aristotelian tradition requires treating people \textit{as individuals} \citep{thiery2006fairness,jorgensen2022algorithms}, not based on the data of others.\footnote{We leave open the question to what extent this can be realized by human decision makers.} For the same reason, what is called \textit{individual fairness} in machine learning fails to be genuinely individual, as pointed out by \citet{binns2020apparent}. Problematically the collective is implicit, hidden, in the mode of personalization; without transparency and explainability, individuals cannot recognize their own context. Insurance scholars have also argued that this diminishes opportunity for collective action \citep{moor2018price,mcfall2018or,krippner2022person} --- the study of collective action in machine learning has just begun \citep{hardt2023algorithmic}.

Another way of framing this consists in problematizing the conceptual foundation of probability and statistics itself. Besides the subjective variant of probability, which we consider unfit for decision making affecting people,\footnote{Setting insurance premia based on subjective probabilities seems objectionable when it affects the welfare of people; similarly for machine learning.} probability and statistics are fundamentally based on aggregates \citep{desrosieres1998politics}. Currently, no viable concept of individual probability is available \citep{dawid2017individual}; instead, probability relies on a reference class \citep{Reichenbach1949-REITTO-3,hajek2007reference}. While multi-calibration aims at finer aggregates, it is still not individual \citep{dawid2017individual}. Thus, even speaking of an `individual probability', which perfect actuarial fairness aims at, has no sound conceptual basis. 
In fact, \citet[Chapter 4]{friedman2020probable} provides an insightful account for the close link of frequentist probability and insurance in the solidaristic mode of the welfare state. This account invites us to consider a reference class as a class of solidarity, which implies disregarding the quest for the single `right' reference class and instead recognizing the normative element in this choice. Furthermore, randomness can then also be viewed as a normative assumption in the face of uncertainty, establishing shared responsibility: ``anyone of us could have had the accident''. Thus, the kind of data that \citet{Venn1876} has in mind, combining aggregate regularity with local irregularity (randomness), corresponds to the prerequisites for insurance. As a consequence, we find normative character in frequentist probability itself and a link to aggregate-based solidarity.
Attempting to individualize frequentist probability then raises a paradox.
In the context of insurance, \citet{frezal2020fairness} have argued that the actuarially fair expected value is only adequate from the economic, aggregate viewpoint of the insurer, but conceptually inadequate (and hence in particular not necessarily `fair') for the individual (see also \citep{frezal2016}). Or, in the words of \citet{abraham1985efficiency}: ``No one has a true expected loss''. When taken to the limit, actuarial fairness thus undermines the logic of insurance itself \citep{martinez2016fair}. In the context of machine learning, we argue that it is problematic to evoke the idea of individual probability (risk), particularly when stakes are high such as in the COMPAS case \citep{machinebias}, even when individual probability is beyond reach and has no conceptual foundation to rest on. We thus encourage more modesty about the epistemic potential of machine learning.



\section{Related Work}
Conceptual literature at the intersection of fair machine learning and insurance is sparse. 
\citet{donahue2021better} study the problem of \textit{externalities of size} by taking inspiration from insurance and challenge the clear distinction between actuarial fairness and solidarity as a consequence. \citet{loi2021choosing} study the interplay of machine learning and (non)discrimination in an insurance setting from a normative, philosophical perspective. \citet{frees2023discriminating} and \citet{charpentier2022} provide broad overviews on discrimination in insurance and consider implications of using machine learning. \citet{xin2022anti} link formal fairness definitions to insurance on a technical level. The closest work to ours that we are aware of is by \citet{barry2022fairness}, who investigate how the use of machine learning in insurance is related to classical fairness debates, but they set other foci. A general difference from ours to related work is that we do not study the use of machine learning \textit{in} insurance, but are interested in a more abstract conceptual linkage. We also note that \citeauthor{frezal2020fairness}'s \citeyearpar{frezal2020fairness} critique of actuarial fairness was a major source of inspiration to us.

\section{Discussion}
The main claim of our work is that insurance is an insightful analogon for the social situatedness and impact of machine learning systems. By traversing this conceptual bridge, machine learning scholars can make use of the rich and interdisciplinary literature on insurance. In particular, we suggest that the multifaceted concept of responsibility, tightly linked to causality and control, deserves more attention. 
We have illustrated problems with actuarial fairness as a notion of fairness in the normative sense. 
In this way, our suggestions are in line with others who demand moving beyond formal fairness to substantive fairness \citep{green2022escaping} and argue that accurate predictive models need not be `fair' \citep{eidelson2021patterned}.

While in this text we have focused on social issues, there are also technical lessons that machine learning could take from insurance, a technology for handling uncertainty. For instance, the problems of \textit{dataset shift} and \textit{model ambiguity} have been recognized in insurance as well as machine learning; for contributions from insurance see \eg \citep{milevsky2006killing,cabantous2007ambiguity,pichler2014insurance,dietz2021pricing}. 
On the other hand, the use of machine learning in insurance is increasing. We thus believe that a research agenda linking machine and learning and insurance may lead to a fruitful, two-way interaction of these fields.

In summary, we offer the following insights. Recent impossibility theorems in the fair machine learning literature \citep{kleinberg_et_al:LIPIcs.ITCS.2017.43} are not as surprising when considering them in the light of the old, fundamental tension in insurance between solidarity and actuarial fairness. In essence, this tension is grounded in how individuals are related to the aggregates they form. This relation rests on responsibilization. Responsibility and responsibilization should be conceptually distinguished, even if in the recent mode of personalized insurance (tightly linked to machine learning) the two are increasingly intertwined. For insurance and machine learning purposes, reasoning about responsibility crucially requires reasoning about causality and control. We have emphasized that the relevant control question has a normative flavour, and thus cannot be left to engineers alone. What is under individuals' control is often hotly contested. As a general research heuristic, many case studies by social scholars of insurance can inspire analogous studies in the context of machine learning, covering a diverse set of topics; mining the literature is thus a rich source of inspiration.


\section*{Acknowledgments}
This work was was funded by the Deutsche Forschungsgemeinschaft (DFG, German Research Foundation)
under Germany’s Excellence Strategy — EXC number 2064/1 — Project number 390727645. The authors
thank the International Max Planck Research School for Intelligent Systems (IMPRS-IS) for supporting
Christian Fröhlich. 
Thanks to Benedikt Höltgen, Sebastian Zezulka, Renate Baumgartner and Maiju Tanninen for helpful discussions and comments. 


\bibliography{insurancebib}

\begin{thebibliography}{138}
\providecommand{\natexlab}[1]{#1}
\providecommand{\url}[1]{\texttt{#1}}
\expandafter\ifx\csname urlstyle\endcsname\relax
  \providecommand{\doi}[1]{doi: #1}\else
  \providecommand{\doi}{doi: \begingroup \urlstyle{rm}\Url}\fi

\bibitem[Abraham(1985)]{abraham1985efficiency}
Kenneth~S. Abraham.
\newblock Efficiency and fairness in insurance risk classification.
\newblock \emph{Virginia Law Review}, 71\penalty0 (3):\penalty0 403--451, 1985.

\bibitem[Andersen \& Nielsen(2015)Andersen and Nielsen]{andersen2015luck}
Martin~Marchman Andersen and Morten Ebbe~Juul Nielsen.
\newblock Luck egalitarianism, universal health care, and
  non-responsibility-based reasons for responsibilization.
\newblock \emph{Res Publica}, 21:\penalty0 201--216, 2015.

\bibitem[Angwin et~al.(2016)Angwin, Larson, Mattu, and Kirchner]{machinebias}
Julia Angwin, Jeff Larson, Surya Mattu, and Lauren Kirchner.
\newblock Machine bias: There’s software used across the country to predict
  future criminals, and it’s biased against blacks.
\newblock \emph{ProPublica}, 2016.
\newblock URL
  \url{https://www.propublica.org/article/machine-bias-risk-assessments-in-criminal-sentencing}.
\newblock Accessed on June 2, 2023.

\bibitem[Armstrong(2005)]{armstrong2005equality}
Chris Armstrong.
\newblock Equality, risk and responsibility: Dworkin on the insurance market.
\newblock \emph{Economy and Society}, 34\penalty0 (3):\penalty0 451--473, 2005.

\bibitem[Arrow(1963)]{arrow1963uncertainty}
Kenneth~J. Arrow.
\newblock Uncertainty and the welfare economics of medical care.
\newblock \emph{The American Economic Review}, 53\penalty0 (5):\penalty0
  941--973, 1963.

\bibitem[Austin(1962)]{austin1962how}
John~L. Austin.
\newblock \emph{How to Do Things With Words}.
\newblock Harvard University Press, Cambridge, 1962.

\bibitem[Avraham(2018)]{avraham2017discrimination}
Ronen Avraham.
\newblock Discrimination and insurance.
\newblock In \emph{The Routledge Handbook of the Ethics of Discrimination}.
  Routledge, 2018.

\bibitem[Avraham et~al.(2013)Avraham, Logue, and
  Schwarcz]{avraham2013understanding}
Ronen Avraham, Kyle~D. Logue, and Daniel Schwarcz.
\newblock Understanding insurance antidiscrimination law.
\newblock \emph{Southern California Law Review}, 87:\penalty0 195--274, 2013.
\newblock URL \url{https://scholarship.law.umn.edu/faculty_articles/576}.
\newblock Accessed on June 2, 2023.

\bibitem[Baker(1996)]{baker1996genealogy}
Tom Baker.
\newblock On the genealogy of moral hazard.
\newblock \emph{Texas Law Review}, 75:\penalty0 237, 1996.

\bibitem[Baker(2000)]{baker2000insuring}
Tom Baker.
\newblock Insuring morality.
\newblock \emph{Economy and {S}ociety}, 29\penalty0 (4):\penalty0 559--577,
  2000.

\bibitem[Baker(2002)]{baker2002risk}
Tom Baker.
\newblock \emph{Risk, insurance, and the social construction of
  responsibility}.
\newblock University of Chicago Press, 2002.

\bibitem[Baker \& Simon(2002)Baker and Simon]{baker2002embracing}
Tom Baker and Jonathan Simon.
\newblock \emph{Embracing risk: The changing culture of insurance and
  responsibility}.
\newblock University of Chicago Press, 2002.

\bibitem[Barocas et~al.(2019)Barocas, Hardt, and
  Narayanan]{barocas-hardt-narayanan}
Solon Barocas, Moritz Hardt, and Arvind Narayanan.
\newblock \emph{Fairness and Machine Learning: Limitations and Opportunities}.
\newblock 2019.
\newblock URL \url{http://www.fairmlbook.org}.
\newblock Accessed on January 5, 2024.

\bibitem[Barry(2019)]{barry2019rationality}
Laurence Barry.
\newblock The rationality of the digital governmentality.
\newblock \emph{Journal for Cultural Research}, 23\penalty0 (4):\penalty0
  365--380, 2019.

\bibitem[Barry(2020)]{barry2020insurance}
Laurence Barry.
\newblock Insurance, big data and changing conceptions of fairness.
\newblock \emph{European Journal of Sociology/Archives Europ{\'e}ennes de
  Sociologie}, 61\penalty0 (2):\penalty0 159--184, 2020.

\bibitem[Barry \& Charpentier(2022)Barry and Charpentier]{barry2022fairness}
Laurence Barry and Arthur Charpentier.
\newblock The fairness of machine learning in insurance: New rags for an old
  man?
\newblock \emph{arXiv preprint arXiv:2205.08112}, 2022.

\bibitem[Binns(2020)]{binns2020apparent}
Reuben Binns.
\newblock On the apparent conflict between individual and group fairness.
\newblock In \emph{Proceedings of the 2020 conference on fairness,
  accountability, and transparency}, pp.\  514--524, 2020.

\bibitem[Bj{\"o}rk et~al.(2020)Bj{\"o}rk, Helgesson, and Juth]{bjork2020better}
Joar Bj{\"o}rk, Gert Helgesson, and Niklas Juth.
\newblock Better in theory than in practise? challenges when applying the luck
  egalitarian ethos in health care policy.
\newblock \emph{Medicine, Health Care and Philosophy}, 23:\penalty0 735--742,
  2020.

\bibitem[Boldyrev \& Svetlova(2016)Boldyrev and Svetlova]{boldyrev2016enacting}
Ivan Boldyrev and Ekaterina Svetlova.
\newblock \emph{Enacting dismal science: New perspectives on the performativity
  of economics}.
\newblock Springer, 2016.

\bibitem[Bowker \& Star(2000)Bowker and Star]{bowker1999sorting}
Geoffrey~C. Bowker and Susan~Leigh Star.
\newblock \emph{Sorting Things Out: Classification and Its Consequences}.
\newblock The MIT Press, 2000.

\bibitem[Brisset(2016)]{brisset2016economics}
Nicolas Brisset.
\newblock Economics is not always performative: some limits for performativity.
\newblock \emph{Journal of Economic Methodology}, 23\penalty0 (2):\penalty0
  160--184, 2016.

\bibitem[Cabantous(2007)]{cabantous2007ambiguity}
Laure Cabantous.
\newblock Ambiguity aversion in the field of insurance: Insurers’ attitude to
  imprecise and conflicting probability estimates.
\newblock \emph{Theory and Decision}, 62\penalty0 (3):\penalty0 219--240, 2007.

\bibitem[Callon(1998)]{callon1998laws}
Michel Callon (ed.).
\newblock \emph{The laws of the markets}.
\newblock Oxford: Blackwell, 1998.

\bibitem[Castelnovo et~al.(2022)Castelnovo, Crupi, Greco, Regoli, Penco, and
  Cosentini]{castelnovo2022clarification}
Alessandro Castelnovo, Riccardo Crupi, Greta Greco, Daniele Regoli,
  Ilaria~Giuseppina Penco, and Andrea~Claudio Cosentini.
\newblock A clarification of the nuances in the fairness metrics landscape.
\newblock \emph{Scientific Reports}, 12, 2022.
\newblock Article number: 4209.

\bibitem[Cevolini \& Esposito(2020)Cevolini and Esposito]{cevolini2020pool}
Alberto Cevolini and Elena Esposito.
\newblock From pool to profile: Social consequences of algorithmic prediction
  in insurance.
\newblock \emph{Big Data \& Society}, 7\penalty0 (2):\penalty0 1--11, 2020.

\bibitem[Cevolini \& Esposito(2022)Cevolini and
  Esposito]{cevolini2022actuarial}
Alberto Cevolini and Elena Esposito.
\newblock From actuarial to behavioural valuation. the impact of telematics on
  motor insurance.
\newblock \emph{Valuation Studies}, 9\penalty0 (1):\penalty0 109--139, 2022.

\bibitem[Charpentier(2022)]{charpentier2022}
Arthur Charpentier.
\newblock Insurance: Discrimination, biases \& fairness.
\newblock \emph{Opinions \& Debates}, 2022.
\newblock URL
  \url{https://www.institutlouisbachelier.org/en/insurance-discrimination-biases-fairness/}.
\newblock Accessed on June 2, 2023.

\bibitem[Clifford \& Iuculano(1987)Clifford and Iuculano]{clifford1987aids}
Karen~A. Clifford and Russel~P. Iuculano.
\newblock {AIDS} and insurance: the rationale for {AIDS}-related testing.
\newblock \emph{Harvard Law Review}, 100\penalty0 (7):\penalty0 1806--1825,
  1987.

\bibitem[Cooper \& Grinder(2009)Cooper and Grinder]{cooper2009probability}
Dan Cooper and Brian Grinder.
\newblock Probability, gambling and the origins of risk management.
\newblock \emph{Financial History Magazine}, 93:\penalty0 10--11, 2009.

\bibitem[D'Amour et~al.(2020)D'Amour, Srinivasan, Atwood, Baljekar, Sculley,
  and Halpern]{d2020fairness}
Alexander D'Amour, Hansa Srinivasan, James Atwood, Pallavi Baljekar, David
  Sculley, and Yoni Halpern.
\newblock Fairness is not static: deeper understanding of long term fairness
  via simulation studies.
\newblock In \emph{Proceedings of the 2020 Conference on Fairness,
  Accountability, and Transparency}, pp.\  525--534, 2020.

\bibitem[Daniels(1990)]{daniels1990insurability}
Norman Daniels.
\newblock Insurability and the {HIV} epidemic: ethical issues in underwriting.
\newblock \emph{The Milbank Quarterly}, 68\penalty0 (4):\penalty0 497--525,
  1990.

\bibitem[Daston(2023)]{daston2023classical}
Lorraine Daston.
\newblock \emph{Classical Probability in the Enlightenment}.
\newblock Princeton University Press, 2023.

\bibitem[Dawid(2017)]{dawid2017individual}
Philip Dawid.
\newblock On individual risk.
\newblock \emph{Synthese}, 194\penalty0 (9):\penalty0 3445--3474, 2017.

\bibitem[de~Finetti(1974/2017)]{de2017theory}
Bruno de~Finetti.
\newblock \emph{Theory of probability: A critical introductory treatment}.
\newblock John Wiley \& Sons, 1974/2017.

\bibitem[Dean(1998)]{dean1998risk}
Mitchell Dean.
\newblock Risk, calculable and incalculable.
\newblock \emph{Soziale Welt}, pp.\  25--42, 1998.

\bibitem[Denuit et~al.(2021)Denuit, Charpentier, and
  Trufin]{denuit2021autocalibration}
Michel Denuit, Arthur Charpentier, and Julien Trufin.
\newblock Autocalibration and {T}weedie-dominance for insurance pricing with
  machine learning.
\newblock \emph{Insurance: Mathematics and Economics}, 101:\penalty0 485--497,
  2021.

\bibitem[Desrosi{\`e}res(1998)]{desrosieres1998politics}
Alain Desrosi{\`e}res.
\newblock \emph{The politics of large numbers: A history of statistical
  reasoning}.
\newblock Harvard University Press, 1998.

\bibitem[Diaz-Bone \& Didier(2016)Diaz-Bone and Didier]{diaz2016introduction}
Rainer Diaz-Bone and Emmanuel Didier.
\newblock Introduction: The sociology of quantification --- perspectives on an
  emerging field in the social sciences.
\newblock \emph{Historical Social Research}, 41\penalty0 (2):\penalty0 7--26,
  2016.

\bibitem[Dietz \& Nieh{\"o}rster(2021)Dietz and
  Nieh{\"o}rster]{dietz2021pricing}
Simon Dietz and Falk Nieh{\"o}rster.
\newblock Pricing ambiguity in catastrophe risk insurance.
\newblock \emph{The Geneva Risk and Insurance Review}, 46\penalty0
  (2):\penalty0 112--132, 2021.

\bibitem[Donahue \& Barocas(2021)Donahue and Barocas]{donahue2021better}
Kate Donahue and Solon Barocas.
\newblock Better together? how externalities of size complicate notions of
  solidarity and actuarial fairness.
\newblock In \emph{Proceedings of the 2021 Conference on Fairness,
  Accountability, and Transparency}, pp.\  185--195, 2021.

\bibitem[Dwork et~al.(2012)Dwork, Hardt, Pitassi, Reingold, and
  Zemel]{dwork2012fairness}
Cynthia Dwork, Moritz Hardt, Toniann Pitassi, Omer Reingold, and Richard Zemel.
\newblock Fairness through awareness.
\newblock In \emph{Proceedings of the 3rd Innovations in Theoretical Computer
  Science Conference}, pp.\  214--226, 2012.

\bibitem[Eidelson(2021)]{eidelson2021patterned}
Benjamin Eidelson.
\newblock Patterned inequality, compounding injustice, and algorithmic
  prediction.
\newblock \emph{American Journal of Law and Equality}, 1:\penalty0 252--276,
  2021.

\bibitem[Ericson et~al.(2000)Ericson, Barry, and Doyle]{ericson2000moral}
Richard Ericson, Dean Barry, and Aaron Doyle.
\newblock The moral hazards of neo-liberalism: lessons from the private
  insurance industry.
\newblock \emph{Economy and Society}, 29\penalty0 (4):\penalty0 532--558, 2000.

\bibitem[Espeland \& Sauder(2007)Espeland and Sauder]{espeland2007rankings}
Wendy~Nelson Espeland and Michael Sauder.
\newblock Rankings and reactivity: How public measures recreate social worlds.
\newblock \emph{American Journal of Sociology}, 113\penalty0 (1):\penalty0
  1--40, 2007.

\bibitem[Espeland \& Stevens(2008)Espeland and Stevens]{espeland2008sociology}
Wendy~Nelson Espeland and Mitchell~L. Stevens.
\newblock A sociology of quantification.
\newblock \emph{European Journal of Sociology/Archives Europ{\'e}ennes de
  Sociologie}, 49\penalty0 (3):\penalty0 401--436, 2008.

\bibitem[Ewald(1989)]{ewald1989versicherungs}
Fran{\c{c}}ois Ewald.
\newblock Die {V}ersicherungs-{G}esellschaft.
\newblock \emph{Kritische Justiz}, 22\penalty0 (4):\penalty0 385--393, 1989.

\bibitem[Ewald(1990)]{ewald1990norms}
Fran{\c{c}}ois Ewald.
\newblock Norms, discipline, and the law.
\newblock \emph{Representations}, 30:\penalty0 138--161, 1990.

\bibitem[Ewald(1991)]{ewald1991insurance}
Francois Ewald.
\newblock Insurance and risk.
\newblock In \emph{The Foucault effect: Studies in governmentality}, pp.\
  197--210. The University of Chicago Press, 1991.

\bibitem[Ewald(1986)]{ewald_letat_1986}
François Ewald.
\newblock \emph{L'État providence}.
\newblock Grasset, 1986.

\bibitem[Frees \& Huang(2023)Frees and Huang]{frees2023discriminating}
Edward~W. Frees and Fei Huang.
\newblock The discriminating (pricing) actuary.
\newblock \emph{North American Actuarial Journal}, 27\penalty0 (1):\penalty0
  2--24, 2023.

\bibitem[Frezal(2016)]{frezal2016}
Sylvestre Frezal.
\newblock Alea and heterogeneity: the tyrannous conflation, 2016.
\newblock URL
  \url{https://www.chaire-pari.fr/wp-content/uploads/2016/09/Alea-and-Heterogeneity_the-Tyrannous-Conflation_eng_MEP.pdf}.
\newblock Accessed on June 14, 2023.

\bibitem[Frezal \& Barry(2020)Frezal and Barry]{frezal2020fairness}
Sylvestre Frezal and Laurence Barry.
\newblock Fairness in uncertainty: Some limits and misinterpretations of
  actuarial fairness.
\newblock \emph{Journal of Business Ethics}, 167:\penalty0 127--136, 2020.

\bibitem[Friedman(2020)]{friedman2020probable}
Rachel~Z. Friedman.
\newblock \emph{Probable Justice: Risk, Insurance, and the Welfare State}.
\newblock University of Chicago Press, 2020.

\bibitem[Gaulding(1994)]{gaulding1994race}
Jill Gaulding.
\newblock Race, sex and genetic discrimination in insurance: What's fair?
\newblock \emph{Cornell Law Review}, 80:\penalty0 1646, 1994.

\bibitem[George~A.(1970)]{george1970market}
Ackerlof George~A.
\newblock The market for lemons: Quality uncertainty and the market mechanism.
\newblock \emph{The Quarterly Journal of Economics}, 84\penalty0 (3):\penalty0
  488--500, 1970.

\bibitem[Gigerenzer et~al.(1989)Gigerenzer, Swijtink, Porter, Daston, and
  Kruger]{gigerenzer1989empire}
Gerd Gigerenzer, Zeno Swijtink, Theodore Porter, Lorraine Daston, and Lorenz
  Kruger.
\newblock \emph{The empire of chance: How probability changed science and
  everyday life}.
\newblock Cambridge University Press, 1989.

\bibitem[Glenn(2003{\natexlab{a}})]{glenn2003postmodernism}
Brian~J. Glenn.
\newblock Postmodernism: the basis of insurance.
\newblock \emph{Risk Management and Insurance Review}, 6\penalty0 (2):\penalty0
  131--143, 2003{\natexlab{a}}.

\bibitem[Glenn(2003{\natexlab{b}})]{glenn2003risk}
Brian~J. Glenn.
\newblock Risk, insurance, and the changing nature of mutual obligation.
\newblock \emph{Law \& Social Inquiry}, 28\penalty0 (1):\penalty0 295--314,
  2003{\natexlab{b}}.

\bibitem[Green(2022)]{green2022escaping}
Ben Green.
\newblock Escaping the impossibility of fairness: From formal to substantive
  algorithmic fairness.
\newblock \emph{Philosophy \& Technology}, 35, 2022.
\newblock Article number: 90.

\bibitem[Gromm{\'e} \& Scheel(2020)Gromm{\'e} and Scheel]{gromme2020doing}
Francisca Gromm{\'e} and Stephan Scheel.
\newblock Doing statistics, enacting the nation: The performative powers of
  categories.
\newblock \emph{Nations and nationalism}, 26\penalty0 (3):\penalty0 576--593,
  2020.

\bibitem[Hafermalz et~al.(2016)Hafermalz, Riemer, and
  Boell]{hafermalz2016enactment}
Ella Hafermalz, Kai Riemer, and Sebastian Boell.
\newblock Enactment or performance? a non-dualist reading of {G}offman.
\newblock In \emph{Beyond Interpretivism? New Encounters with Technology and
  Organization: IFIP WG 8.2 Working Conference on Information Systems and
  Organizations, IS\&O 2016}, pp.\  167--181. Springer, Cham, 2016.

\bibitem[H{\'a}jek(2007)]{hajek2007reference}
Alan H{\'a}jek.
\newblock The reference class problem is your problem too.
\newblock \emph{Synthese}, 156:\penalty0 563--585, 2007.

\bibitem[Hardt et~al.(2022)Hardt, Jagadeesan, and
  Mendler-D{\"u}nner]{hardt2022performative}
Moritz Hardt, Meena Jagadeesan, and Celestine Mendler-D{\"u}nner.
\newblock Performative power.
\newblock In \emph{Advances in Neural Information Processing Systems},
  volume~35, pp.\  22969--22981, 2022.

\bibitem[Hardt et~al.(2023)Hardt, Mazumdar, Mendler-D\"{u}nner, and
  Zrnic]{hardt2023algorithmic}
Moritz Hardt, Eric Mazumdar, Celestine Mendler-D\"{u}nner, and Tijana Zrnic.
\newblock Algorithmic collective action in machine learning.
\newblock In \emph{Proceedings of the 40th International Conference on Machine
  Learning}, ICML'23. 2023.

\bibitem[Hasson et~al.(2020)Hasson, Nastase, and Goldstein]{hasson2020direct}
Uri Hasson, Samuel~A. Nastase, and Ariel Goldstein.
\newblock Direct fit to nature: An evolutionary perspective on biological and
  artificial neural networks.
\newblock \emph{Neuron}, 105\penalty0 (3):\penalty0 416--434, 2020.

\bibitem[Heidari et~al.(2019)Heidari, Loi, Gummadi, and
  Krause]{heidari2019moral}
Hoda Heidari, Michele Loi, Krishna~P. Gummadi, and Andreas Krause.
\newblock A moral framework for understanding fair {ML} through economic models
  of equality of opportunity.
\newblock In \emph{Proceedings of the 2019 Conference on Fairness,
  Accountability, and Transparency}, pp.\  181--190, 2019.

\bibitem[Heimer(1985)]{heimer1985reactive}
Carol~Anne Heimer.
\newblock \emph{Reactive risk and rational action: Managing moral hazard in
  insurance contracts}.
\newblock University of California Press, 1985.

\bibitem[Heras et~al.(2020)Heras, Pradier, and Teira]{martinez2016fair}
Antonio~J. Heras, Pierre-Charles Pradier, and David Teira.
\newblock What was fair in actuarial fairness?
\newblock \emph{History of the Human Sciences}, 33\penalty0 (2):\penalty0
  91--114, 2020.

\bibitem[H{\"o}ltgen \& Williamson(2023)H{\"o}ltgen and
  Williamson]{holtgen2023richness}
Benedikt H{\"o}ltgen and Robert~C. Williamson.
\newblock On the richness of calibration.
\newblock In \emph{Proceedings of the 2023 Conference on Fairness,
  Accountability, and Transparency}, pp.\  1124–1138. 2023.

\bibitem[Hu \& Chen(2018)Hu and Chen]{hu2018short}
Lily Hu and Yiling Chen.
\newblock A short-term intervention for long-term fairness in the labor market.
\newblock In \emph{Proceedings of the 2018 World Wide Web Conference}, pp.\
  1389--1398, 2018.

\bibitem[Hu \& Kohler-Hausmann(2020)Hu and Kohler-Hausmann]{hu2020s}
Lily Hu and Issa Kohler-Hausmann.
\newblock What's sex got to do with machine learning?
\newblock In \emph{Proceedings of the 2020 Conference on Fairness,
  Accountability, and Transparency}, pp.\  513, 2020.

\bibitem[Huseby(2016)]{huseby2016can}
Robert Huseby.
\newblock Can luck egalitarianism justify the fact that some are worse off than
  others?
\newblock \emph{Journal of Applied Philosophy}, 33\penalty0 (3):\penalty0
  259--269, 2016.

\bibitem[Jorgensen(2022)]{jorgensen2022algorithms}
Ren{\'e}e Jorgensen.
\newblock Algorithms and the individual in criminal law.
\newblock \emph{Canadian Journal of Philosophy}, 52\penalty0 (1):\penalty0
  61--77, 2022.

\bibitem[Joseph et~al.(2016)Joseph, Kearns, Morgenstern, and
  Roth]{joseph2016fairness}
Matthew Joseph, Michael Kearns, Jamie~H. Morgenstern, and Aaron Roth.
\newblock Fairness in learning: Classic and contextual bandits.
\newblock In \emph{Advances in Neural Information Processing Systems},
  volume~29. 2016.

\bibitem[Kasirzadeh(2022)]{kasirzadeh2022algorithmic}
Atoosa Kasirzadeh.
\newblock Algorithmic fairness and structural injustice: Insights from feminist
  political philosophy.
\newblock In \emph{Proceedings of the 2022 AAAI/ACM Conference on AI, Ethics,
  and Society}, AIES '22, pp.\  349–356, New York, NY, USA, 2022. Association
  for Computing Machinery.
\newblock ISBN 9781450392471.

\bibitem[Kilbertus et~al.(2017)Kilbertus, Rojas~Carulla, Parascandolo, Hardt,
  Janzing, and Sch{\"o}lkopf]{kilbertus2017avoiding}
Niki Kilbertus, Mateo Rojas~Carulla, Giambattista Parascandolo, Moritz Hardt,
  Dominik Janzing, and Bernhard Sch{\"o}lkopf.
\newblock Avoiding discrimination through causal reasoning.
\newblock In \emph{Advances in neural information processing systems},
  volume~30, 2017.

\bibitem[Kleinberg et~al.(2017)Kleinberg, Mullainathan, and
  Raghavan]{kleinberg_et_al:LIPIcs.ITCS.2017.43}
Jon Kleinberg, Sendhil Mullainathan, and Manish Raghavan.
\newblock {Inherent Trade-Offs in the Fair Determination of Risk Scores}.
\newblock In \emph{8th Innovations in Theoretical Computer Science Conference
  (ITCS 2017)}, volume~67, pp.\  43:1--43:23, 2017.

\bibitem[Knight(2013)]{knight2013luck}
Carl Knight.
\newblock Luck egalitarianism.
\newblock \emph{Philosophy Compass}, 8\penalty0 (10):\penalty0 924--934, 2013.

\bibitem[Krippner(2023)]{krippner2023unmasked}
Greta~R. Krippner.
\newblock Unmasked: A history of the individualization of risk.
\newblock \emph{Sociological Theory}, pp.\  83--104, 2023.

\bibitem[Krippner \& Hirschman(2022)Krippner and Hirschman]{krippner2022person}
Greta~R. Krippner and Daniel Hirschman.
\newblock The person of the category: the pricing of risk and the politics of
  classification in insurance and credit.
\newblock \emph{Theory and Society}, 51\penalty0 (5):\penalty0 685--727, 2022.

\bibitem[Kuppler et~al.(2021)Kuppler, Kern, Bach, and
  Kreuter]{kuppler2021distributive}
Matthias Kuppler, Christoph Kern, Ruben~L Bach, and Frauke Kreuter.
\newblock Distributive justice and fairness metrics in automated
  decision-making: How much overlap is there?
\newblock \emph{arXiv preprint arXiv:2105.01441}, 2021.

\bibitem[Landes(2013)]{landes2013normative}
Xavier Landes.
\newblock The normative foundations of (social) insurance: Technology, social
  practices and political philosophy.
\newblock 2013.
\newblock URL
  \url{https://www.centroeinaudi.it/images/abook_file/WP-LPF_6_2013_Landes.pdf}.
\newblock Accessed on June 14, 2023.

\bibitem[Landes(2015)]{landes2015fair}
Xavier Landes.
\newblock How fair is actuarial fairness?
\newblock \emph{Journal of Business Ethics}, 128:\penalty0 519--533, 2015.

\bibitem[Landes \& Holtug(2015)Landes and Holtug]{landes2015insurance}
Xavier Landes and Nils Holtug.
\newblock Insurance, equality and the welfare state: Political philosophy and
  (of) public insurance.
\newblock \emph{Res Publica}, 21:\penalty0 111--118, 2015.

\bibitem[Lee(1993)]{lee1993racial}
Sharon~M. Lee.
\newblock Racial classifications in the {US} census: 1890--1990.
\newblock \emph{Ethnic and Racial Studies}, 16\penalty0 (1):\penalty0 75--94,
  1993.

\bibitem[Lehtonen \& Liukko(2011)Lehtonen and Liukko]{lehtonen2011forms}
Turo-Kimmo Lehtonen and Jyri Liukko.
\newblock The forms and limits of insurance solidarity.
\newblock \emph{Journal of Business Ethics}, 103:\penalty0 33--44, 2011.

\bibitem[Lehtonen \& Liukko(2015)Lehtonen and Liukko]{lehtonen2015producing}
Turo-Kimmo Lehtonen and Jyri Liukko.
\newblock Producing solidarity, inequality and exclusion through insurance.
\newblock \emph{Res publica}, 21\penalty0 (2):\penalty0 155--169, 2015.

\bibitem[Lehtonen \& Van~Hoyweghen(2014)Lehtonen and
  Van~Hoyweghen]{lehtonen2014editorial}
Turo-Kimmo Lehtonen and Ine Van~Hoyweghen.
\newblock Editorial: Insurance and the economization of uncertainty journal.
\newblock \emph{Journal of Cultural Economy}, 7\penalty0 (4):\penalty0
  532--540, 2014.

\bibitem[Lippert-Rasmussen(2015)]{lippert2015genetic}
Kasper Lippert-Rasmussen.
\newblock Genetic discrimination and health insurance.
\newblock \emph{Res Publica}, 21\penalty0 (2):\penalty0 185--199, 2015.

\bibitem[Liu et~al.(2018)Liu, Dean, Rolf, Simchowitz, and
  Hardt]{liu2018delayed}
Lydia~T. Liu, Sarah Dean, Esther Rolf, Max Simchowitz, and Moritz Hardt.
\newblock Delayed impact of fair machine learning.
\newblock In \emph{International Conference on Machine Learning}, pp.\
  3150--3158, 2018.

\bibitem[Liukko(2010)]{liukko2010genetic}
Jyri Liukko.
\newblock Genetic discrimination, insurance, and solidarity: an analysis of the
  argumentation for fair risk classification.
\newblock \emph{New Genetics and Society}, 29\penalty0 (4):\penalty0 457--475,
  2010.

\bibitem[Loi \& Christen(2021)Loi and Christen]{loi2021choosing}
Michele Loi and Markus Christen.
\newblock Choosing how to discriminate: Navigating ethical trade-offs in fair
  algorithmic design for the insurance sector.
\newblock \emph{Philosophy \& Technology}, 34:\penalty0 967--992, 2021.

\bibitem[Lupton(2016)]{lupton2016diverse}
Deborah Lupton.
\newblock The diverse domains of quantified selves: self-tracking modes and
  dataveillance.
\newblock \emph{Economy and Society}, 45\penalty0 (1):\penalty0 101--122, 2016.

\bibitem[MacKenzie(2008)]{mackenzie2008engine}
Donald MacKenzie.
\newblock \emph{An engine, not a camera: How financial models shape markets}.
\newblock {MIT} Press, 2008.

\bibitem[MacKenzie et~al.(2008)MacKenzie, Muniesa, and
  Siu]{mackenzie2007economists}
Donald MacKenzie, Fabian Muniesa, and Leung-Sea Siu (eds.).
\newblock \emph{Do economists make markets?: on the performativity of
  economics}.
\newblock Princeton University Press, 2008.

\bibitem[M{\"a}ki(2013)]{maki2013performativity}
Uskali M{\"a}ki.
\newblock Performativity: Saving austin from mackenzie.
\newblock In \emph{EPSA11 perspectives and foundational problems in philosophy
  of science}, pp.\  443--453, 2013.

\bibitem[McFall(2011)]{mcfall2011good}
Liz McFall.
\newblock A ‘good, average man’: Calculation and the limits of statistics
  in enrolling insurance customers.
\newblock \emph{The Sociological Review}, 59\penalty0 (4):\penalty0 661--684,
  2011.

\bibitem[McFall(2019)]{mcfall2019personalizing}
Liz McFall.
\newblock Personalizing solidarity? the role of self-tracking in health
  insurance pricing.
\newblock \emph{Economy and Society}, 48\penalty0 (1):\penalty0 52--76, 2019.

\bibitem[McFall \& Moor(2018)McFall and Moor]{mcfall2018or}
Liz McFall and Liz Moor.
\newblock Who, or what, is insurtech personalizing?: persons, prices and the
  historical classifications of risk.
\newblock \emph{Distinktion: journal of social theory}, 19\penalty0
  (2):\penalty0 193--213, 2018.

\bibitem[McFall et~al.(2020)McFall, Meyers, and
  Hoyweghen]{mcfall2020personalisation}
Liz McFall, Gert Meyers, and Ine~Van Hoyweghen.
\newblock Editorial: The personalisation of insurance: Data, behaviour and
  innovation.
\newblock \emph{Big Data \& Society}, 7\penalty0 (2):\penalty0 1--11, 2020.

\bibitem[Mehrabi et~al.(2021)Mehrabi, Morstatter, Saxena, Lerman, and
  Galstyan]{mehrabi2021survey}
Ninareh Mehrabi, Fred Morstatter, Nripsuta Saxena, Kristina Lerman, and Aram
  Galstyan.
\newblock A survey on bias and fairness in machine learning.
\newblock \emph{ACM Computing Surveys (CSUR)}, 54\penalty0 (6):\penalty0 1--35,
  2021.

\bibitem[Mennicken \& Espeland(2019)Mennicken and Espeland]{mennicken2019s}
Andrea Mennicken and Wendy~Nelson Espeland.
\newblock What's new with numbers? sociological approaches to the study of
  quantification.
\newblock \emph{Annual Review of Sociology}, 45:\penalty0 223--245, 2019.

\bibitem[Meyers \& Van~Hoyweghen(2018)Meyers and
  Van~Hoyweghen]{meyers2018enacting}
Gert Meyers and Ine Van~Hoyweghen.
\newblock Enacting actuarial fairness in insurance: From fair discrimination to
  behaviour-based fairness.
\newblock \emph{Science as Culture}, 27\penalty0 (4):\penalty0 413--438, 2018.

\bibitem[Milevsky et~al.(2006)Milevsky, Promislow, and
  Young]{milevsky2006killing}
M.~A. Milevsky, S.~D. Promislow, and V.~R. Young.
\newblock Killing the law of large numbers: Mortality risk premiums and the
  sharpe ratio.
\newblock \emph{Journal of Risk and Insurance}, 73\penalty0 (4):\penalty0
  673--686, 2006.

\bibitem[Miller(2009)]{miller2009disparate}
Michael~J. Miller.
\newblock Disparate impact and unfairly discriminatory insurance rates.
\newblock In \emph{Casualty Actuarial Society E-Forum, Winter 2009}, 2009.

\bibitem[Mitchell et~al.(2021)Mitchell, Potash, Barocas, D'Amour, and
  Lum]{mitchell2021algorithmic}
Shira Mitchell, Eric Potash, Solon Barocas, Alexander D'Amour, and Kristian
  Lum.
\newblock Algorithmic fairness: Choices, assumptions, and definitions.
\newblock \emph{Annual Review of Statistics and Its Application}, 8:\penalty0
  141--163, 2021.

\bibitem[Mol(2002)]{mol2002body}
Annemarie Mol.
\newblock \emph{The body multiple: Ontology in medical practice}.
\newblock Duke University Press, 2002.

\bibitem[Moor \& Lury(2018)Moor and Lury]{moor2018price}
Liz Moor and Celia Lury.
\newblock Price and the person: Markets, discrimination, and personhood.
\newblock \emph{Journal of Cultural Economy}, 11\penalty0 (6):\penalty0
  501--513, 2018.

\bibitem[Mora(2014)]{mora2014making}
G.~Cristina Mora.
\newblock \emph{Making Hispanics: How Activists, Bureaucrats, and Media
  Constructed a New American}.
\newblock University of Chicago Press, 2014.

\bibitem[Perdomo et~al.(2020)Perdomo, Zrnic, Mendler-D{\"u}nner, and
  Hardt]{perdomo2020performative}
Juan Perdomo, Tijana Zrnic, Celestine Mendler-D{\"u}nner, and Moritz Hardt.
\newblock Performative prediction.
\newblock In \emph{International Conference on Machine Learning}, pp.\
  7599--7609, 2020.

\bibitem[Pichler(2014)]{pichler2014insurance}
Alois Pichler.
\newblock Insurance pricing under ambiguity.
\newblock \emph{European Actuarial Journal}, 4\penalty0 (2):\penalty0 335--364,
  2014.

\bibitem[Prainsack \& Van~Hoyweghen(2020)Prainsack and
  Van~Hoyweghen]{prainsack2020shifting}
Barbara Prainsack and Ine Van~Hoyweghen.
\newblock Shifting solidarities: Personalisation in insurance and medicine.
\newblock \emph{Shifting solidarities: Trends and developments in European
  societies}, pp.\  127--151, 2020.

\bibitem[Pyysi{\"a}inen et~al.(2017)Pyysi{\"a}inen, Halpin, and
  Guilfoyle]{pyysiainen2017neoliberal}
Jarkko Pyysi{\"a}inen, Darren Halpin, and Andrew Guilfoyle.
\newblock Neoliberal governance and ‘responsibilization’ of agents:
  reassessing the mechanisms of responsibility-shift in neoliberal discursive
  environments.
\newblock \emph{Distinktion: Journal of Social Theory}, 18\penalty0
  (2):\penalty0 215--235, 2017.

\bibitem[R{\"a}z(2021)]{raz2021group}
Tim R{\"a}z.
\newblock Group fairness: Independence revisited.
\newblock In \emph{Proceedings of the 2021 Conference on Fairness,
  Accountability, and Transparency}, pp.\  129--137, 2021.

\bibitem[Rebert \& Van~Hoyweghen(2015)Rebert and
  Van~Hoyweghen]{rebert2015right}
Lisa Rebert and Ine Van~Hoyweghen.
\newblock The right to underwrite gender: The goods \& services directive and
  the politics of insurance pricing.
\newblock \emph{Tijdschrift Voor Genderstudies}, 18\penalty0 (4):\penalty0
  413--431, 2015.

\bibitem[Rechfeld(2016)]{rechfeld2016personalised}
Florian Rechfeld.
\newblock Personalised genetic testing and its impact to insurance.
\newblock \emph{Swiss Re}, 2016.
\newblock URL
  \url{https://www.swissre.com/dam/jcr:24995a5d-5b66-42ea-a2b9-660458bc6e26/Personalised_genetic_testing_and_its_impact_to_insurance.pdf}.
\newblock Accessed on June 14, 2023.

\bibitem[Reichenbach(1949)]{Reichenbach1949-REITTO-3}
Hans Reichenbach.
\newblock \emph{The Theory of Probability: An Inquiry Into the Logical and
  Mathematical Foundations of the Calculus of Probability}.
\newblock University of California Press, 1949.

\bibitem[Rosanvallon(2000)]{Rosanvallon2000-ROSTNS-2}
Pierre Rosanvallon.
\newblock \emph{The New Social Question: Rethinking the Welfare State}.
\newblock Princeton University Press, 2000.

\bibitem[Schw{\"o}bel \& Remmers(2022)Schw{\"o}bel and
  Remmers]{schwobel2022long}
Pola Schw{\"o}bel and Peter Remmers.
\newblock The long arc of fairness: Formalisations and ethical discourse.
\newblock In \emph{Proceedings of the 2022 Conference on Fairness,
  Accountability, and Transparency}, pp.\  2179--2188, 2022.

\bibitem[Selbst et~al.(2019)Selbst, Boyd, Friedler, Venkatasubramanian, and
  Vertesi]{sociotechnical}
Andrew~D. Selbst, Danah Boyd, Sorelle~A. Friedler, Suresh Venkatasubramanian,
  and Janet Vertesi.
\newblock Fairness and abstraction in sociotechnical systems.
\newblock In \emph{Proceedings of the 2019 Conference on Fairness,
  Accountability, and Transparency}, pp.\  59–68. Association for Computing
  Machinery, 2019.

\bibitem[Shamir(2008)]{shamir2008age}
Ronen Shamir.
\newblock The age of responsibilization: On market-embedded morality.
\newblock \emph{Economy and Society}, 37\penalty0 (1):\penalty0 1--19, 2008.

\bibitem[Sharon(2017)]{sharon2017self}
Tamar Sharon.
\newblock Self-tracking for health and the quantified self: Re-articulating
  autonomy, solidarity, and authenticity in an age of personalized healthcare.
\newblock \emph{Philosophy \& Technology}, 30\penalty0 (1):\penalty0 93--121,
  2017.

\bibitem[Stone(2001)]{stone2001admissions}
Deborah~A. Stone.
\newblock Ad missions.
\newblock The American Prospect, 2001.
\newblock URL \url{https://prospect.org/health/ad-missions/}.
\newblock Accessed on May 17, 2023.

\bibitem[Swedloff(2014)]{swedloff2014risk}
Rick Swedloff.
\newblock Risk classification's big data (r) evolution.
\newblock \emph{Connecticut Insurance Law Journal}, 143:\penalty0 339--373,
  2014.

\bibitem[Tanninen(2020)]{tanninen2020contested}
Maiju Tanninen.
\newblock Contested technology: Social scientific perspectives of
  behaviour-based insurance.
\newblock \emph{Big Data \& Society}, 7\penalty0 (2):\penalty0 1--14, 2020.

\bibitem[Thiery \& Van~Schoubroeck(2006)Thiery and
  Van~Schoubroeck]{thiery2006fairness}
Yves Thiery and Caroline Van~Schoubroeck.
\newblock Fairness and equality in insurance classification.
\newblock \emph{The Geneva Papers on Risk and Insurance-Issues and Practice},
  31\penalty0 (2):\penalty0 190--211, 2006.

\bibitem[Van~Hoyweghen(2014)]{van2014politics}
Ine Van~Hoyweghen.
\newblock On the politics of calculative devices: performing life insurance
  markets.
\newblock \emph{Journal of Cultural Economy}, 7\penalty0 (3):\penalty0
  334--352, 2014.

\bibitem[Van~Hoyweghen(2018)]{van2018genomics}
Ine Van~Hoyweghen.
\newblock Genomics and insurance: The lock-in effects of a politics of genetic
  solidarity.
\newblock In \emph{Handbook of Genomics, Health and Society}, pp.\  203--211.
  Routledge, 2018.

\bibitem[Van~Hoyweghen et~al.(2006)Van~Hoyweghen, Horstman, and
  Schepers]{van2006making}
Ine Van~Hoyweghen, Klasien Horstman, and Rita Schepers.
\newblock Making the normal deviant: The introduction of predictive medicine in
  life insurance.
\newblock \emph{Social Science \& Medicine}, 63\penalty0 (5):\penalty0
  1225--1235, 2006.

\bibitem[Van~Hoyweghen et~al.(2007)Van~Hoyweghen, Horstman, and
  Schepers]{van2007genetic}
Ine Van~Hoyweghen, Klasien Horstman, and Rita Schepers.
\newblock Genetic ‘risk carriers’ and lifestyle ‘risk takers’. which
  risks deserve our legal protection in insurance?
\newblock \emph{Health Care Analysis}, 15:\penalty0 179--193, 2007.

\bibitem[Venn(1876)]{Venn1876}
John Venn.
\newblock \emph{The Logic of Chance}.
\newblock MacMillan, 1876.

\bibitem[Verbelen et~al.(2018)Verbelen, Antonio, and
  Claeskens]{verbelen2018unravelling}
Roel Verbelen, Katrien Antonio, and Gerda Claeskens.
\newblock Unravelling the predictive power of telematics data in car insurance
  pricing.
\newblock \emph{Journal of the Royal Statistical Society: Series C (Applied
  Statistics)}, 67\penalty0 (5):\penalty0 1275--1304, 2018.

\bibitem[Vosselman(2014)]{vosselman2014performativity}
Ed~Vosselman.
\newblock The ‘performativity thesis’ and its critics: Towards a relational
  ontology of management accounting.
\newblock \emph{Accounting and Business Research}, 44\penalty0 (2):\penalty0
  181--203, 2014.

\bibitem[Vredenburgh(2022)]{vredenburghfairness}
Kate Vredenburgh.
\newblock {Fairness}.
\newblock In \emph{{The Oxford Handbook of AI Governance}}. Oxford University
  Press, 2022.

\bibitem[Walters(1981)]{walters1981risk}
Michael~A. Walters.
\newblock Risk classification standards.
\newblock In \emph{Proceedings of the Casualty Actuarial Society}, volume~68,
  pp.\  1--18, 1981.

\bibitem[Wilkie(1997)]{wilkie1997mutuality}
David Wilkie.
\newblock Mutuality and solidarity: assessing risks and sharing losses.
\newblock \emph{Philosophical Transactions of the Royal Society of London.
  Series B: Biological Sciences}, 352\penalty0 (1357):\penalty0 1039--1044,
  1997.

\bibitem[Williamson(2004)]{williamson2004dynamic}
Jon Williamson.
\newblock A dynamic interaction between machine learning and the philosophy of
  science.
\newblock \emph{Minds and Machines}, 14\penalty0 (4):\penalty0 539--549, 2004.

\bibitem[Xin \& Huang(2023)Xin and Huang]{xin2022anti}
Xi~Xin and Fei Huang.
\newblock Antidiscrimination insurance pricing: Regulations, fairness criteria,
  and models.
\newblock \emph{North American Actuarial Journal}, pp.\  1--35, 2023.

\end{thebibliography}
\bibliographystyle{tmlr}

\appendix
\section{Appendix}
\section{Non-Responsibility Based Reasons for Responsibilization}
\label{sec:moralhazardadverse}
 The difference between responsibility and responsibilization can be subtle. In insurance, two classical principles supply reasons for justifying responsibilization \textit{without} being based on responsibility, however. We might also call them ``efficiency-based'' reasons for responsibilization \citep{andersen2015luck}.

\textit{Adverse selection} \citep{george1970market,thiery2006fairness,avraham2013understanding} refers to the informational asymmetry between insurer and insured (policyholder) at the time of underwriting. Typically, the insured is better informed about their risk; higher-risk individuals are more likely to seek insurance for protection. The reasoning is then that without segmentation, insurers are more likely to attract high-risk individuals who profit from the subsidy of the pool; ultimately, leading to bankruptcy of the insurer. Thus competition drives increasing segmentation. Public insurance has the benefit of compulsory participation, so that adverse selection cannot occur and solidarity can be implemented. In contrast, adverse selection is advanced by the industry as a justification for actuarial fairness \citep{miller2009disparate}.

\textit{Moral hazard} \citep{heimer1985reactive,baker1996genealogy,baker2000insuring} (in the context of insurance) on the other hand refers to a \textit{performative}, behaviour-shaping aspect of insurance premia. Simply put, the idea is that policyholders are more inclined to behave in a risky way due to the protection offered by insurance coverage. For instance, there is less incentive to purchase precautionary measures such as alarm systems.
Like adverse selection, moral hazard is invoked as an argument in favor of actuarial fairness. The welfare state is seen as the ultimate source of moral hazard \citep{ericson2000moral}. For instance, public health insurance might lead to more visits to the doctor.\footnote{Noting the negative connotation in the term \textit{moral hazard}, \citet{baker1996genealogy} asks: ``What, after all, is wrong
with enabling people to go to the doctor when they feel the need, and why
should we be concerned when they do so?''}
Indeed, moral hazard is \textit{the} responsibilization force in the neoliberal era \citep{ericson2000moral}. In particular, moral hazard favors behaviour-based personalization \citep{verbelen2018unravelling}. Insurance companies thus increasingly emphasize loss prevention, that is, acting on behaviour \citep{baker2002embracing,cevolini2022actuarial}.
Note that the applicability of moral hazard presupposes control for a feedback loop to exist.


Both adverse selection and moral hazard concern the performative dimensions of ``calculative devices'' \citep{van2014politics}. Hence we suggest that these concepts might be usefully transposed onto machine learning.


\section{Proof of Proposition~\ref{prop:actfair}}
\label{app:actfairproof}
\begin{proof}
   
Plugging in the definition of the actuarially fair predictor, we find that $\forall G \in \mathcal{G}$:
        \begin{align}
        \label{eq:mustbecondexp}
        &\E[\hat{Y}(X)|X \in G]=\E[Y|X \in G]\\
\Leftrightarrow  &\E[(x \mapsto \E[Y|X \in G_x])(X)|X \in G]=\E[Y|X \in G]\\
\Leftrightarrow &\E[Y|X \in G] = \E[Y|X \in G],
\end{align}
which is true. For the second statement, observe that Equation~\ref{eq:mustbecondexp} implies that a predictor $\hat{Y}$ which is constant on each group $G$ must equal the conditional expectation for that group.
\end{proof}

\subsection{Independence with respect to all groups implies full solidarity}
\label{app:solidarityproof}
\begin{proposition}
    Assume independence holds with respect to all groups, that is
    \[
    \hat{Y} \indep \chi_A, \quad \forall A \subseteq \Omega, A \neq \emptyset.
    \]
    Then $\forall \omega \in \Omega : P(\{\omega\})>0 : \hat{Y}(x)=c $\thinspace for some $c \in \mathbb{R}$.
\end{proposition}
\begin{proof}
    Recall that we assume a finite $\Omega$. Then we can assume without loss of generality that $P(\{\omega\})>0$ $\forall \omega \in \Omega$, otherwise we could work on an altered probability space by discarding sets of measure zero. From the independence assumption it follows that $P(\{\omega \colon \hat{Y}(\omega) = y | A)=P(\{\omega \colon \hat{Y}(\omega)=y)$ for any $y \in \mathbb{R}$ and $A \subseteq \Omega$, $A \neq \emptyset$. Pick any $\omega_1$ so that $\hat{Y}(\omega_1)=y_1$. Then it must hold $P(\{\omega \colon \hat{Y}(\omega)=y_1|\{\omega_1\}) = 1 = P(\{\omega \colon \hat{Y}(\omega)=y_1\})$, from which we conclude that $\hat{Y}$ must be constant on $\Omega$.
\end{proof}

\section{Performativity}
\label{sec:performativity}
A central and unifying theme that emerges when considering social issues surrounding insurance, and as we contend, also machine learning, is that of \textit{performativity}. This concept allows us to comprehend many of the previously raised points through a new lense. The idea of performativity originates from John L.\@\xspace Austin's seminal work ``How to do things with words'' \citep{austin1962how}, where Austin notes that the function of language is often not only descriptive, but also has constitutive and causal effects. For example, uttering ``I promise'' itself \textit{constitutes} a promise and has the causal effect of establishing certain expectations. \citet{austin1962how} refers to sentences with such \textit{performative} force as \textit{speech acts}. Since Austin, the term `performativity' has travelled far into multiple disciplines and acquired lives of its own, so that it can be hard to pin down precisely a common core (see \citep{maki2013performativity} for a critique). We will use the term in the broadest possible sense to emphasize similarity of perspectives instead of differences.
An influential account has been put forward in economics \citep{callon1998laws,mackenzie2007economists,mackenzie2008engine}, which has inspired also a recent formalization in machine learning \citep{perdomo2020performative,hardt2022performative}. The central claim of this line of work is that economics is not simply in the business of describing or representing an independent, passive reality, but also actively shapes it, for instance by encouraging people to act in accordance with its models \citep{boldyrev2016enacting}. Another general framework and a source of inspiration to us, can be found in the work of \citet{mol2002body}. To avoid the dualist connotation of the term performance, implying a `backstage reality',\footnote{However, for a critical examination of whether this dualism is inherently associated with `performance', see \citep{hafermalz2016enactment}.} \citet{mol2002body} instead coins the term \textit{enactment}, referring to the multiple and ongoing work that sustains a reality:
\begin{quote}
It is possible to say that in practices objects are \textit{enacted}. This suggests that activities take place --- but
leaves the actors vague. It also suggests that in the act, and only then and there,
something \textit{is} --- being enacted. [emphasis in original]
\end{quote}
Perhaps the simplest way to understand what is at the heart of performativity, we suggest, is to assert that \textit{representation and intervention are entangled} \citep{vosselman2014performativity}. 
Performativity hence contrasts with the commonsense view that perception and action can be neatly separated; in the latter, the task of machine learning is simply to extract patterns from a passive reality `out there' in an objective way.
For instance, \citet{mitchell2021algorithmic} distinguish between ``world as it is'' and ``world as it should and could be'', mapping onto prediction and decision task. Similarly, \citet{kuppler2021distributive} urge to cleanly separate prediction and decision.
A word that frequently occurs in this context is `bias', which we like to avoid, since we associate it with the notion of an objective `backstage' reality, whose representation is then distorted. 

What then is the relevance of performativity for insurance, and by analogy, machine learning? Actuarial fairness (calibration), or more broadly the fairness of `accurate' statistical methods, carries with it an aura of objectivity and neutrality. \textit{If} we choose responsibilization, \textit{then} our predictive models better be `objective and neutral'. Recall that actuarial fairness aims to set premia in accordance with the expected risk for each policyholder, and it is assumed that insurers can know this risk through statistical methods. \citet{glenn2003postmodernism} succinctly captures this as follows:
\begin{quote}
    [T]here is a general belief that
insurance practices are predicated on objective statistics, what has elsewhere been called
“the myth of the actuary”. The myth of the actuary is the idea that there
is a reality in the world that can be captured by rational choice models and statistical
analysis—and that insurance companies do this ethically, objectively, and “correctly.”
\end{quote}
Such objectivity then is supposed to be a source of authority and fairness. In the words of \citet{van2014politics}:
\begin{quote}
    The dominant view is that
insurance technologies of risk assessment are somehow ‘measuring’, ‘observing’ or
‘describing’ peoples’ insurance risks. This paper calls for a different approach, namely a
pragmatist analysis of the \textit{performativity of insurance calculative devices}. Contrary to the
financial realism of the everyday categories of insurance numbers, I argue that insurance
calculative devices not only represent but generate, intervene and rearrange the worlds in
which they are deployed. [emphasis added]
\end{quote}

The performativity of insurance and machine learning becomes especially relevant due to ethical implications. Many scholars have argued, providing insightful examples, that insurance is fundamentally a \textit{normative technology}  \citep{baker2002embracing,glenn2003postmodernism,van2006making,van2007genetic,lehtonen2015producing,tanninen2020contested,prainsack2020shifting}, depending on causality, control and responsibility. \textit{Doing} insurance or machine learning involves \textit{enacting} certain realities and suppressing others, as we have sketched in Section~\ref{sec:contingencies}. For instance, in the process of collecting data, only some features are considered, and others neglected. Expanding on this, a performativity perspective would emphasize that there is no objective data `collection' process, that quantification and categorization require significant and ongoing work; such work may be influenced by implicit normative judgements, which becomes ingrained and hidden in the `representation'. There is now a vibrant, if still nascent, research field on the \textit{sociology of quantification} (including categorization), owing much to the seminal work of \citet{desrosieres1998politics}; for overviews of this field see \citep{espeland2008sociology,diaz2016introduction,mennicken2019s}, where the reader finds plenty of evidence for such work. Central in this research field is again performativity, or what has been called the \textit{constitutive potential} of quantification  \citep{mennicken2019s}. 
As a noteworthy example, it has been demonstrated that the census, through the introduction of statistical categories, can contribute to the establishment of a collective identity among the individuals it aims to describe \citep{lee1993racial,bowker1999sorting,mora2014making}. Thus, a category that was initially intended to merely represent acquires performativity by actively shaping the formation of this particular group \citep{gromme2020doing}.\footnote{To anticipate a criticism, this does of course not imply that arbitrary sets of people can become a mutually recognizing group: the hard conceptual work is to investigate how performativity of groups functions and what its limits are. 
In Austinian terms, this means delineating the ``felicity'' conditions, which make performative utterances successful \citep{brisset2016economics}.
} 

We propose that insurance can act as a model for the performativity of statistical, ``calculative devices'' \citep{van2014politics} that arise from their social situatedness. In machine learning, performativity shows up in at least two ways: on the one hand, it requires training data, and this training data has been shaped by performative forces in a broader social context --- referring to the quantification and categorization processes. Using the data thus imports this performativity into the model. On the other hand, by deploying a machine learning model its predictions may acquire performative force in both a constitutive and causal sense. The predictions may act as interventions and through responsibilization shape the behaviour of people, who for instance may strategically adapt to the predictions --- this sense of performativity is also referred to as \textit{reactivity} \citep{espeland2007rankings,espeland2008sociology}, and is closer to the formalization proposed by \citet{perdomo2020performative}, which emphasizes the causal dimension, but neglects the constitutive one. The extent of this phenomenon, \ie how much a company can steer a population using a model, has been termed \textit{performative power} \citep{hardt2022performative}.

As explicated in Section~\ref{sec:fairmllink}, depending on the choice of fairness metric (or none) and to which features it is applied (\ie the choice of groups), machine learning can exert responsibilizing and non-responsibilizing force. We suggest that two classes of performative effects can then be broadly distinguished, which is however not a clear dichotomy in light of Section~\ref{sec:contingencies}. When machine learning responsibilizes for a controllable feature, individuals may adapt to the prediction so as to change it --- if they receive feedback; this is the hope of personalized insurance, and this setting also motivates the concept of moral hazard (Appendix~\ref{sec:moralhazardadverse}). In contrast, blindly applying the principle of actuarial fairness can lead to responsibilizing for non-controllable features, which then runs the risk of reproducing past injustice implicit in the training data. Yet against a background of such past injustice, it is not clear why actuarial \textit{fairness} should be considered as a principle of \textit{justice} --- this argument has been made both in the insurance \citep{daniels1990insurability,lehtonen2015producing,barry2020insurance} as well as the machine learning literature \citep{mitchell2021algorithmic,vredenburghfairness,green2022escaping,kasirzadeh2022algorithmic}; see also \citep{eidelson2021patterned} --- however, the insurance literature provides illuminating examples. In this way, machine learning (resp.\@\xspace insurance) can implicitly responsibilize for sensitive features such as gender or race; the situation is particularly intricate when a feature is considered as controllable which stands in a constitutive relation to a sensitive feature \citep{hu2020s}, for instance due to performativity.

In response to the performativity of machine learning, we advocate for explicit reflection about how performative forces have shaped the present input data, and furthermore how a model in conjunction with a choice of fairness metric might exert performative force by acting on people. Focusing on (non)responsibilization and performativity implies taking a dynamic perspective. Thus, it becomes imperative to foreground and explicitly model the effects of deploying machine learning systems
 \citep{hu2018short,liu2018delayed,d2020fairness,schwobel2022long}, constrasting with rather static vocabulary such as \textit{bias} or \textit{discrimination}.
In this process of reflexive inquiry, we suggest to pay more attention to enactments of causality, control and responsibility --- framing them in this way rather than as immutable \textit{facts} implies making them contestable, that is, putting them on the stage for scrutiny \citep{glenn2003postmodernism}. 

\end{document}